\documentclass[preprint,10pt,authoryear]{elsarticle}

\usepackage{amsmath, amsfonts, amssymb, bm}

\usepackage[english]{babel}
\usepackage{geometry} % 页面边距
\usepackage{booktabs, tabularx, adjustbox, multirow, makecell}

\usepackage{graphicx, wrapfig, subfigure}
\usepackage{tikz}
\usepackage{pgfplots}
\usepackage{subcaption} % 支持 \begin{subfigure}	
\pgfplotsset{compat=1.18}

\usepackage{algorithm, algorithmicx, algpseudocode}

\usepackage{listings, color}

\usepackage{hyperref}
\usepackage{url}

\usepackage{lipsum} % 占位符文本

\definecolor{MyPurple}{RGB}{128, 0, 128}
\definecolor{mypink}{rgb}{0.93, 0.01, 0.55}
\definecolor{myblue}{rgb}{0.149, 0.541, 0.851}
\definecolor{mygreen}{rgb}{0.416, 0.8, 0.408}
\definecolor{myteal}{RGB}{0, 128, 128}
\usepackage{sectsty}
\allsectionsfont{\normalfont\bfseries}

\begin{document}
		
	\begin{frontmatter}
		\title{KOSS: Kalman-Optimal Selective State Spaces for Long-Term Sequence Modeling}
		
		\author[addr1]{Lei Wang}
		\author[addr1]{Xin Tan\corref{cor1}}
		\author[addr1]{Mingwei Wang}
		\author[addr1]{Ying Zhang}
		
		\address[addr1]{School of Electronic Information and Artificial Intelligence, Shaanxi University of Science and Technology}
		\cortext[cor1]{Corresponding author.}
		
		\begin{abstract}
			Recent selective state space models (SSMs), such as Mamba and Mamba-2, have demonstrated strong performance in sequence modeling owing to input-dependent selection mechanisms. However, these mechanisms lack theoretical grounding and cannot support context-aware selection from latent state dynamics. To address these limitations, we propose KOSS: a Kalman-optimal Selective State Space model that formulates selection as latent state uncertainty minimization. Derived from estimation theory, KOSS adopts a continuous-time latent update: $\hat{h}'(t) = A_K \hat{h}(t) + B_K x(t) + K x'(t)$, where the Kalman gain $K$  dynamically modulates information propagation based on content and context. This enables a closed-loop, context-aware selectivity mechanism. To ensure stable computation and near-linear scalability, KOSS employs global spectral differentiation for frequency-domain derivative estimation, along with a segment-wise scan for hardware-efficient processing. On a selective copying task with distractors, KOSS achieves over 79\% accuracy while baselines drop below 20\%, demonstrating robust context-aware selection. Furthermore, across nine long-term forecasting benchmarks, KOSS reduces MSE by 2.92–36.23\% and consistently outperforms state-of-the-art models in both accuracy and stability. To assess real-world applicability, a case study on secondary surveillance radar (SSR) tracking confirms KOSS’s robustness under irregular intervals and noisy conditions and demonstrates its effectiveness in real-world applications. Finally, supplementary experiments verify Kalman gain convergence and the frequency response of spectral differentiation, providing theoretical support for the proposed closed-loop design.
		\end{abstract}
		
	\end{frontmatter}

	\section{Introduction}
	Long-term time series forecasting (LTSF) aims to predict future values over extended horizons based on past time series values.. It is critical in domains such as energy scheduling, climate modeling, finance, and traffic management, where accurate long-range forecasts inform strategic decisions and resource allocation. Compared to short-term forecasting, LTSF presents unique challenges: it requires modeling long-range dependencies, mitigating error accumulation, and handling complex and often non-stationary temporal dynamics~\cite{cheng2025comprehensive}.
	To address these challenges, a range of methods have been proposed. Recurrent Neural Networks (RNNs) are widely used to capture temporal dependencies~\cite{salazar2025distance, benjamin2025enhancing}, but their performance on long sequences is often hindered by vanishing or exploding gradients~\cite{pascanu2013difficulty}. Transformers~\cite{vaswani2017attention}, based on attention mechanisms~\cite{bahdanau2014neural}, have shown strong empirical performance in modeling long-range dependencies by flexibly routing information across positions. However, their reliance on fixed-length attention windows imposes limitations in extrapolating beyond the observed context, and their quadratic complexity in sequence length poses scalability challenges in practice.
	
	\vspace{0.1em}
	\noindent
	Recently, selective state space models (SSMs)~\cite{gu2023mamba, dao2024transformers} have emerged as a compelling alternative for sequence modeling. These models dynamically adjust SSM parameters based on the input content, enabling them to filter task-irrelevant patterns while maintaining linear-time complexity through fused scan operations on modern hardware. Mamba and its variants have achieved strong performance in domains such as language~\cite{bondaschi2025markov}, audio~\cite{lee2025deft}, and genomics~\cite{wu2025generator}. However, their robustness in LTSF remains limited: when exposed to long-range dependencies, anomalous inputs, or distribution shifts, these models often degrade due to overfitting to short-term or input-correlated patterns.
	We identify two critical limitations in existing selective SSMs. 
	\textbf{\textit{First, the lack of a theoretical foundation for selection:}} current mechanisms are dependent on input-driven feedforward neural networks, lack both an explicit optimization objective and theoretical guarantees for their selective behavior. 
	\textbf{\textit{Second, inability to perform context-based reasoning:}} existing selection mechanisms generate dynamic parameters solely from current input. This input-only perspective prevents adaptation to contextual history.
	
	We propose \textbf{\textit{Kalman-Based Optimization of Selective State Space Models (KOSS)}}, a novel class of selective SSMs grounded in the Kalman optimal estimation principle~\cite{kalman1960new}.
	
	\paragraph{Kalman-Optimal SSM.}
	We begin by formulating selection as the problem of minimizing latent state uncertainty. Starting from the principle of minimum mean squared estimation error~\cite{wan2025information}, we derive a continuous-time SSM with Kalman-optimal properties:
	\(\hat{h}'(t) = {A_K}\hat{h}(t) + {B_K}x(t) + Kx'(t)\) 
	where \(K\) is the Kalman gain~\cite{kalman1960new}, and the dynamics \(A_K\) and \(B_K\) are expressed as linear transformations of \(K\).
	
	\paragraph{Innovation-Driven Selectivity (IDS).}
	We design a novel class of selection mechanism by parameterizing the Kalman-Optimal SSM dynamics using the \textit{innovation} (i.e., the discrepancy between the input and historical state).  This enables the model to selectively filter out irrelevant information and remember relevant information over long horizons, based on both the input (content) and the latent state (context).
	
	\paragraph{Spectral Differentiation Unit (SDU).}
	To enable stable and accurate estimation of input derivatives in the Kalman-Optimal SSM, we propose the \textit{Spectral Differentiation Unit (SDU)}, which performs derivative estimation in the frequency domain by leveraging the global receptive field of Fourier transforms. This spectral approach overcomes the instability and noise sensitivity inherent to local difference methods, making it better suited for long-sequence modeling while ensuring stable state propagation and efficient gradient-based learning.
	
	\paragraph{Segment-wise Parallel Scan.}
	The dynamic coupling between model parameters and latent states, which is introduced by our innovation-driven selectivity mechanism, breaks the assumptions behind existing hardware-friendly architectures. To preserve linear-time scalability under this coupling, we propose a hardware-aware\textit{Segment-wise Parallel Scan} scheme that parallelizes computation within segments while propagating recurrent dependencies across segment boundaries. This design strikes a balance between modeling fidelity and hardware efficiency, and adapts flexibly to diverse deployment scenarios through tunable segment lengths.
	
	\vspace{1em}
	\noindent
	We validate KOSS through comprehensive experiments across synthetic and real-world forecasting tasks. On the Selective Copying task with correlated distractors, KOSS retains over 79\% accuracy under 50\% interference. Across nine long-term forecasting benchmarks, KOSS achieves 10\%–30\% lower MSE than state-of-the-art baselines and uniquely sustains accuracy as prediction horizons increase. Dynamic scalability analysis further reveals that KOSS matches Mamba’s throughput when the segment length is approximately half of the total sequence length, while achieving higher accuracy at smaller $S$.
	Additionally, we evaluate KOSS using raw plot data from a field-deployed Secondary Surveillance Radar (SSR)~\cite{barbary2021industrial}, where KOSS demonstrates superior robustness in handling noisy, sparse, and irregular real-world time series—validating its generalization under authentic operational constraints.

	\section{Related Work: State Space Models}
	Structured (S4) and Selective (S6) State Space Models represent a recent class of deep sequence models that are closely related to RNNs, CNNs, and classical state space models. They are inspired by a specific continuous-time system ~(\ref{eq:SSM}) that maps a one-dimensional input function or sequence \( x(t) \in \mathbb{R} \) to an output \( y(t) \in \mathbb{R} \) through an implicit latent state \( h(t) \in \mathbb{R}^N \).
	
	Concretely, both S4 and S6 models are specified by four parameters \((\Delta, A, B, C)\), which define a sequence-to-sequence transformation in two stages:
	\begin{equation}
		\begin{aligned}
			h'(t) &= A h(t) + B x(t) \\
			y(t) &= C h(t)
		\end{aligned}
		\label{eq:SSM}
	\end{equation}
	
	\begin{equation}
		\begin{aligned}
			h_{t} &= \Bar{A} h_{t-1} + \Bar{B} x_{t} \\
			y_{t} &= C h_{t}
		\end{aligned}
		\label{eq:Discretization}
	\end{equation}

	\subsection{Discretization}
	In the first stage, the continuous-time parameters \((\Delta, A, B)\) are converted into discrete-time parameters \((\Bar{A}, \Bar{B})\) via fixed formulas \(\Bar{A} = f_A(\Delta, A)\) and \(\Bar{B} = f_B(\Delta, A, B)\), where the function pair \((f_A, f_B)\) is commonly referred to as a discretization rule. A typical example is the zero-order hold (ZOH) defined in equation~(\ref{eq:ZOH}):

	\begin{equation}
		\Bar{A} = \exp(\Delta A) \qquad \Bar{B} = (\Delta A)^{-1} (\exp(\Delta A) - I) \cdot \Delta B
		\label{eq:ZOH}
	\end{equation}
	
	Discretization plays a critical role in bridging continuous-time formulations and their implementation as discrete computation graphs in deep SSMs. Specifically, it inherits beneficial properties from the underlying continuous dynamics, such as resolution invariance~\cite{nguyen2022s4nd} and automatic normalization~\cite{gu2022train, orvieto2023resurrecting}, which are crucial for stable training and generalization. From a computational standpoint, discretization constitutes the first operation in the forward pass of an SSM. Some recent SSM variants~\cite{zhang2023effectively} bypass the discretization process entirely by directly parameterizing the discrete-time matrices \((\Bar{A}, \Bar{B})\), offering a more direct and interpretable modeling approach.
	
	\subsection{Selective Mechanism} \label{sec:Related_Work_selective_mechanisms}
	
	\paragraph{Selection as a Means of Compression}
	In sequence models, the trade-off between computational efficiency and representational effectiveness is fundamentally governed by the ability to compress state representations: efficient models rely on compact states to reduce computation, while effective models must have a state that contains all necessary information from the context. Recent work~\cite{gu2023mamba} formulates this compression process as a problem of selectivity, defined as the context-aware ability to focus on or filter out inputs during the construction of sequential states. In particular, a selection mechanism determines how information is propagated and interacts across time.

	\begin{table}[!htbp]
		\centering
		\renewcommand{\arraystretch}{1.6}
		\setlength{\tabcolsep}{10pt}
		\begin{tabular}{@{}>{\centering\arraybackslash}p{0.5cm} 
				>{\centering\arraybackslash}p{4.2cm} 
				>{\centering\arraybackslash}p{4.2cm} 
				>{\centering\arraybackslash}p{4.9cm}@{}}
			\toprule
			\makecell{\textbf{}} 
			& \makecell{\textbf{S4(Structural)}} 
			& \makecell{\textbf{S6 (Input-Dependent)}} 
			& \makecell{\textbf{KOSS(Context-Aware)}} \\
			\midrule
			
			\makecell{$\Delta$ }
			& \makecell{$\tau_{\Delta}(\text{Parameter}) \rightarrow \Delta \in \mathbb{R}^{D}$} 
			& \makecell{$\tau_{\Delta}(\text{Parameter + $\mathcal{S}_\Delta(x)$}) \rightarrow$ \\ 
				$\Delta \in \mathbb{R}^{B \times L \times D}$} 
			& \makecell{\textbf{\boldmath$\tau_{\Delta}(\text{Parameter + $\mathcal{S}_\Delta(x)$}) \rightarrow$} \\ 
				\textbf{\boldmath$\Delta \in \mathbb{R}^{B \times L \times D}$}} \\
			
			\addlinespace
			\makecell{$A$}
			& \makecell{$\text{Parameter} \rightarrow A \in \mathbb{R}^{D \times N};$ \\
				$A_0 = \text{HiPPO}$} 
			& \makecell{$\text{Parameter} \rightarrow A \in \mathbb{R}^{D \times N};$ \\
				$\bar{A} = \exp(\Delta A)$} 
			& \makecell{\textbf{\boldmath$\text{Parameter} \rightarrow A \in \mathbb{R}^{D \times N};$} \\
				\textbf{\boldmath$\mathcal{F}_A(A,\phi(\mathit{Innov},C)) \rightarrow A_K$}} \\
			
			\addlinespace
			\makecell{$B$}
			& \makecell{$\text{Parameter} \rightarrow B \in \mathbb{R}^{D \times N}$}
			& \makecell{	$\mathcal{S}_B(x) \rightarrow B \in \mathbb{R}^{B \times L \times D}$} 
			& \bm{$\mathcal{F}_B(A,\phi(\mathit{Innov},C) \rightarrow {B_K}$} \\
			
			\addlinespace
			\makecell{$C$} 
			& \makecell{$\mathcal{S}_C(x) \rightarrow C \in \mathbb{R}^{B \times L \times D}$}
			& \makecell{$\mathcal{S}_C(x) \rightarrow C \in \mathbb{R}^{B \times L \times D}$}
			& \bm{$\mathcal{S}_C(x) \rightarrow C \in \mathbb{R}^{B \times L \times D}$}\\
			
			\bottomrule
		\end{tabular}
		\caption{
			This comparison highlights a fundamental shift from static, structure-induced selectivity in S4 to dynamic, input-dependent mechanisms in S6. For reference, we also include our proposed KOSS, which introduces a new paradigm of context-aware, innovation-dependent selectivity. A detailed discussion of KOSS will be provided in Chapter~\ref{sec:selective_mechanisms}.}
		\label{tab:table_selective_mechanisms} % Add label for referencing the table
	\end{table}
	
	\paragraph{Classic Selection Mechanism}
	Early structured SSMs such as S4 implement such mechanisms implicitly by encoding fixed inductive biases into their parameterized linear dynamics (e.g., HiPPO-initialized $A$ matrices)~\cite{gu2020hippo}, resulting in uniform filtering over time without input-dependent adaptation. Motivated by synthetic tasks such as selective copy and induction heads, S6 proposes an explicit, input-dependent selection mechanism that parameterizes the SSM parameters \((\Delta, B, C)\) based on the input. This allows the model to filter out irrelevant information and remember relevant information indefinitely, based on input content and temporal location.
	
	\vspace{0.5em}
	\noindent
	We summarize the evolution of these mechanisms in Table~\ref{tab:table_selective_mechanisms}, comparing how S4, S6, and our proposed KOSS implement selectivity in key components of their state-space structures. For a deeper discussion of the theoretical background and historical development of selection mechanisms, we refer the reader to Appendix~\ref{sec:Appendix_Selection_Mechanism}.

	\subsection{Linear Time-Varying System}
	As shown in Table~\ref{tab:table_selective_mechanisms}, S4 and S6 illustrate two representative selection mechanisms. The main difference is simply making several parameters \((A, B, C)\) functions of the input, which introduces corresponding changes in tensor shapes. In particular, these parameters now have a length dimension \(L\), meaning that the model has changed from time-invariant to time-varying.
	
	\paragraph{Computation}
	To maintain efficiency under time-varying dynamics, recent works such as Mamba~\cite{gu2023mamba} adopt a hardware-aware scan algorithm for efficient parallelizable training, and switch to linear recurrent mode for efficient autoregressive inference.

	\paragraph{LTV Dynamics}
	While LTV systems break the constraints of fixed state propagation paths and thus offer more flexible and expressive modeling of long-range dependencies, a core insight of our work is that purely input-dependent LTV models face inherent limitations when modeling certain types of data. We provide a detailed analysis of this limitation in Section~\ref{sec:Motivation}, which motivates our subsequent design that removes such LTV constraints while overcoming the associated efficiency bottlenecks.
	
	\vspace{0.5em}
	\noindent
	This transition not only improves modeling capacity but also motivates the use of scan-style computation in modern selective SSMs. A more detailed discussion of this transition is provided in Appendix~\ref{appendix:Appendix_LTV}.

	\section{Kalman-Based Selective SSM} \label{sec:selective_mechanisms}
	This section introduces a Kalman-based selective mechanism that addresses the limitations of input-only selection in prior SSMs. Inspired by Kalman filtering and minimum mean square error (MMSE) estimation, we formulate a context-aware selection strategy that jointly considers current inputs and latent states. The resulting model provides a theoretically grounded approach to dynamic information routing, connecting selective SSMs with attention-like behaviors.
	
	\subsection{Motivation: Towards Context-Aware Selectivity} \label{sec:Motivation}
	We concur with the observation that a fundamental principle in sequence modeling is to compress contextual information into a smaller state via selective mechanisms. We further emphasize that the key to effective compression lies in the \textbf{\textit{precision and efficiency}} of these selective mechanisms, which depend not only on the input but also on the historical context embedded in latent states. This perspective is exemplified in Transformer models, where attention weights—viewed as dynamic selection coefficients—are computed through interactions between the current token (query) and the full contextual sequence (keys and values), enabling context-aware selection across time. Notably, the parameter matrices in SSM (e.g., \((A,B,C,D)\)) can be interpreted as selection coefficients akin to attention weights, governing how information is routed across time. In models such as S4, these coefficients are fixed, while in S6, they are dynamically modulated by the input. Both lack the ability to condition selection on historical context accumulated from previous computations.
	
\begin{figure}[!htbp]
		\centering
		\includegraphics[width=\textwidth]{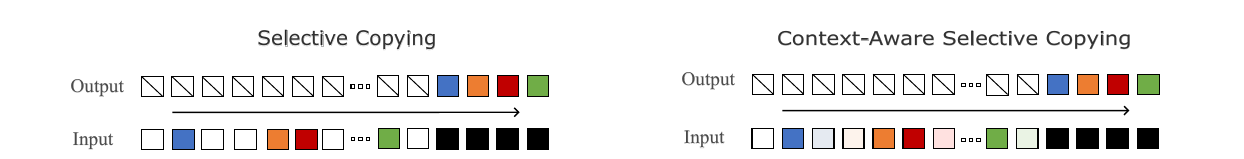}
		\caption{
			(Left) The Selective Copying task introduces random spacing between inputs and output elements, and can be effectively solved by input-dependent LTV models. 
			(Right) The Context-Aware Selective Copying task extends this setup by adding faint-colored correlated distractors, and requires time-varying models that can precisely and selectively remember or ignore inputs depending on their content and historical context.
		}
		\label{fig:copy-vs-selective}
	\end{figure}

	To understand this principle, we revisit the classic synthetic task of the \textbf{\textit{Selective Copying}}~\cite{gu2023mamba}, and propose a synthetic yet illustrative variant Figure~\ref{fig:copy-vs-selective}.
	
	\paragraph{Context-Aware Selective Copying}task introduces distractors that emulate correlated noise frequently observed in real-world sequences, which are inputs that superficially resemble relevant patterns but no longer carry task-relevant information. To solve the task, the model must retain truly relevant tokens (dark-colored) while identifying and filtering out both irrelevant (white) and misleading (faint-colored) ones. This setup directly challenges the model’s ability to perform context-aware selection, where selection weights should depend not only on the current input but also on historical latent states.
	
	These synthetic tasks are very similar to the Denoising task~\cite{jing2019gated}. The Selective Copying task is a variant of the popular Copying task~\cite{arjovsky2016unitary}, characterized by variable spacing between inputs-to-outputs~(Figure~\ref{fig:copy-vs-selective} Left). This requires the model’s dynamics to be input-aware. As a result, static global convolutional modeling methods fail on this task, while input-dependent LTV models can easily solve it by detecting whether the input is non-zero. 
	However, when the inputs contains distractors that are correlated with target patterns~(Figure~\ref{fig:copy-vs-selective}, Right), such input-dependent mechanisms begin to fail, as the Context-Aware Selective Copying task requires the ability to focus on or ignore inputs based not only on their content, but also on the historical context compressed into the sequential state.
	
	In summary, the effectiveness of sequence models depends on their ability to efficiently and accurately compress context into a compact state. In turn, we propose context-aware selectivity: a selection process jointly governed by the input and the accumulated latent states. We formalize this concept in Section~\ref{sec:Kalman-Optimal Continuous SSMs} and ~\ref{sec:Improving SSMs} via a Kalman-optimal framework designed to achieve optimal information filtering under semantic uncertainty.
	
	\subsection{Kalman-Optimal SSM}\label{sec:Kalman-Optimal Continuous SSMs}
	As established in Section~\ref{sec:Motivation}, an effective context-aware selective mechanism should dynamically determine whether, and to what extent, to focus on or filter out inputs, based on both their content and the accumulated context in latent states. Inspired by classical estimation theory~\cite{kalman1960new}, we cast the selective behavior of compressing inputs into a latent state as an optimal estimation problem, which provides a principled framework for understanding how information propagates and interacts along the sequence dimension. 
	
	In particular, the recurrent dynamics of LTV SSMs can be restructured under a Kalman-inspired framework, where the system parameters (e.g., $A$, $B$) are dynamically optimized using a gain term computed from the \textit{innovation}~\cite{ghaleb2017improved}. This \textit{innovation}-driven adjustment enables the model to adaptively control its dynamics based on content (input) and context (hidden states), achieving selective information routing with Kalman-optimal properties. We call this formulation the Kalman-Optimal Selective Mechanism.

	\paragraph{Derive Kalman Gain} In the classical continuous-time Kalman filter, the Kalman gain \( K(t) \) is typically obtained by solving a matrix Riccati differential equation, which describes the evolution of the estimation error covariance. However, in this section, we instead derive the optimal gain functionally, by directly minimizing the instantaneous expected squared estimation error \( \mathbb{E}[\| h(t) - \hat{h}(t) \|^2] \), without explicitly solving the Riccati equation.
	
	Let \( e(t) = h(t) - \hat{h}(t) \) be the estimation error after measurement update. The objective is to minimize the expected squared error:
	\begin{equation}
		J(K(t)) = \mathbb{E}[e(t)^\top e(t)].
	\end{equation}
	Using the standard measurement update rule in Kalman filtering, the posterior estimate \( \hat{h}(t) \) is given by
	\begin{equation}
		\hat{h}(t) = \hat{h}^-(t) + K(t)(x(t) - C \hat{h}^-(t)),
	\end{equation}
	where \( \hat{h}^-(t) \) denotes the prior (a priori) state estimate, i.e., the predicted state before incorporating the current observation. The term \( x(t) - C \hat{h}^-(t) \), known as the \textit{innovation}, quantifies the discrepancy between the  observation and the predicted state. The observation \( x(t) \) follows a linear model:
	
	\begin{equation}
		x(t) = C h(t) + v,
		\label{eq:observation model}
	\end{equation}
	with \( v \) representing the observation noise, assumed to have covariance \( R \).
	
	We can express the estimation error as:
	\begin{equation}
		e(t) = h(t) - \hat{h}(t) = (I - K(t) C)(h(t) - \hat{h}^-(t)) - K(t) v,
		\label{eq:estimation error}
	\end{equation}
	where \( e^-(t) = h(t) - \hat{h}^-(t) \) denotes the prior estimation error (also known as the prediction error).
	
	The covariance of the updated error becomes:
	\begin{equation}
		P(t) = \mathbb{E}[e(t) e(t)^\top] = (I - K(t) C) P^-(t) (I - K(t) C)^\top + K(t) R K(t)^\top,
		\label{eq:update covariance}
	\end{equation}
	where \( P^-(t) \) is the prior error covariance. Minimizing the expected error energy is equivalent to minimizing the trace:
	\begin{equation}
		J(K(t)) = \mathrm{Tr}(P(t)).
	\end{equation}
	
	Taking the derivative with respect to \( K(t) \) and setting it to zero yields the optimal Kalman gain:
	\begin{equation}
		K(t) = P^-(t) C^\top (C P^-(t) C^\top + R)^{-1}.
		\label{eq:Kalman gain}
	\end{equation}
	
	This gain minimizes the posterior uncertainty and thus defines the theoretically optimal way to fuse the current observation with prior belief.
	
	\paragraph{A Continuous-Time SSM}
	To derive a continuous-time state-space model that inherits the optimal estimation characteristics of the Kalman filter, we begin with the classical measurement update equation:
	
	\begin{equation}
		\hat{h}(t) = \hat{h}^-(t) + K(t)(x(t) - C \hat{h}^-(t)),
	\end{equation}
	we take the derivative with respect to time:
	\begin{align}
		\frac{d}{dt} \hat{h}(t) &= \frac{d}{dt} \hat{h}^-(t) + \dot{K}(t)(x - C\hat{h}^-(t)) \nonumber \\
		&\quad + K(t)\left( \frac{d}{dt}x - C\frac{d}{dt}\hat{h}^-(t) \right).
	\end{align}
	
	Assuming the prior follows linear dynamics:
	\begin{equation}
		\frac{d}{dt} \hat{h}^-(t) = A \hat{h}^-(t),
	\end{equation}
	we substitute and express everything in terms of $\hat{h}(t)$ using:
	\begin{equation}
		\hat{h}^-(t) = \hat{h}(t) - K(t)(x - C\hat{h}(t)).
	\end{equation}
	
	These algebraic manipulations yield a continuous-time SSM that embodies Kalman-optimal estimation through a dynamic, feedback-driven formulation:
	\begin{equation}
		\frac{d}{dt} \hat{h}(t) =  A(t) \hat{h}(t) + B(t) x(t) + K(t) \frac{d}{dt} x(t),
		\label{eq:continuous-ssm}
	\end{equation}
	where:
	\begin{align}
		A(t) &= (A - K(t) C A) (I + K(t) C) - \dot{K(t)} C - \dot{K(t)} C K(t) C, \\
		B(t) &= - (A - K(t) C A) K(t) + \dot{K(t)} + \dot{K(t)} C K(t).
	\end{align}

	We adopt the approximation \(\dot{K}(t) \approx 0\) to simplify the model, which is justified by the fact that, under standard observability conditions, the Kalman gain \(K(t)\) provably converges to a steady state. Detailed discussions and empirical validation of this convergence are provided in Appendix~\ref{appendix:kalman-derivative} (Figure~\ref{fig:kalman-gain-convergence}).
	
	The simplified continuous-time state-space formulation is:
	\begin{equation}
		\begin{aligned}
			\frac{d}{dt} \hat{h}(t) &=  A(t) \hat{h}(t) + B(t) x(t) + K(t) \frac{d}{dt} x(t), \\
			A(t) &= (A - K(t) C A)(I + K(t) C), \\
			B(t) &= - (A - K(t) C A) K(t).
		\end{aligned}
		\label{eq:structured-ode}
	\end{equation}
	and enables more stable and efficient neural state-space models for long-range prediction tasks.
	
	\subsection{IDS: Innovation-Driven Selectivity} \label{sec:Improving SSMs}
	Recent work in sequential modeling has begun to incorporate selectivity mechanisms into SSMs. A representative example is Mamba~\cite{gu2023mamba}, which explicitly formulates this principle:
	\begin{quote}
		“One method of incorporating a selection mechanism into models is by letting their parameters that affect interactions along the sequence (e.g., the recurrent dynamics of an RNN or the convolution kernel of a CNN) be input-dependent.”
	\end{quote}
	
	Motivated by this idea, we design an efficient and principled selectivity mechanism within the structured ODE framework of Equation~\ref{eq:structured-ode}. Specifically, we leverage the \textit{innovation}, which inherently encapsulates information about both the latent state and the current input, to dynamically modulate the system parameters in a state-aware manner. In this way, the model retains Kalman-style optimality while introducing dynamic, context-aware control over information flow, as motivated in Section~\ref{sec:Motivation}. This approach differs from models such as Mamba and Mamba-2, where the selectivity mechanism is solely input-dependent without incorporating the latent states.

	\paragraph{Learnable Gain from Innovation} 
	In Kalman filtering, the gain \( K(t) \) depends on the state prediction error \( h(t) - \hat{h}^-(t) \), as shown in Equations~\ref{eq:estimation error}, \ref{eq:update covariance}, \ref{eq:Kalman gain}. 
	Under the linear observation model~(\ref{eq:observation model}), the prediction error in the latent space is mapped into the observation space, yielding the following expression for the \textit{innovation} (denoted by \(\mathit{Innov}\)):
	\begin{equation}
		\mathit{Innov} = x(t) - C \hat{h}^-(t) = C\big(h(t) - \hat{h}^-(t)\big) + v,
	\end{equation}
	
	This formulation reveals that the \textit{innovation} is a linear transformation of the state prediction error, perturbed by additive observation noise \( v\). Consequently, the \textit{innovation} encapsulates the same underlying prediction error information, up to the distortion introduced by the observation process. This relationship provides a theoretical justification for estimating the gain \( K(t) \) directly from the \textit{innovation}, rather than computing it via the classical Kalman update formula:
	\begin{equation}
		K(t) = \phi\left( \mathit{Innov} \right),
	\end{equation}
	where \( \phi(\cdot) \) denotes a generic nonlinear mapping that projects the  \textit{innovation} to the gain space.

	\paragraph{Selectivity via Gain}
	In the structured ODE defined by Equation~\ref{eq:structured-ode}, the system matrices \( A(t) \) and \( B(t) \) are linear transformations of the gain \( K(t) \). To make this functional dependence explicit, we rewrite them as:
	\begin{equation}
		A_K(t) = \mathcal{F}_A(A, K(t), C), \quad
		B_K(t) = \mathcal{F}_B(A, K(t), C),
	\end{equation}
	where the subscript \( K \) is introduced to emphasize that both matrices are dynamically modulated by the gain \( K(t) \), in contrast to fixed system matrices.
	
	With these definitions in place, we now propose a context-aware selectivity mechanism: the gain \( K(t) \) is estimated via a nonlinear mapping from the \textit{innovation}, enabling dynamic, state-dependent modulation. This estimated gain then parameterizes the system matrices \( A_K(t) \) and \( B_K(t) \) through the mappings \( \mathcal{F}_A(\cdot) \) and \( \mathcal{F}_B(\cdot) \), thereby modulating model’s dynamics in a feedback-driven manner.
	As a result, the model can selectively remember or forget information based on both the content (input) and context (hidden states)—unlike prior approaches such as Mamba, which rely solely on input-driven modulation without latent state feedback. 
	
	For a deeper understanding of the Kalman gain as a content-based and context-sensitive selection factor, we provide a detailed reinterpretation, where its role as a theoretically grounded prototype for dynamic selectivity in sequence modeling is discussed (see Appendix~\ref{appendix:Gain is a Selective}).

	\subsection{Spectral Differentiation Unit: From Finite Differences to Frequency-Domain Estimation}\label{sec:SDU}
	
	A key term in our Kalman-Optimal State Space formulation is the input derivative $\frac{d}{dt} x(t)$, which determines how rapidly the input is evolving and guides the dynamic modulation of the gain matrix $K(t)$. In practice, input signals are discrete and noisy, making stable and accurate derivative estimation a non-trivial problem.
	
	\paragraph{Limitations of Finite Differences.}
	Traditional numerical approaches estimate derivatives using finite difference schemes~\cite{vidale1988finite}. For a discrete sequence $x = \{x_0, x_1, \dots, x_{N-1}\}$, the first-order approximation is given by:
	
	\begin{equation}
	x'_n \approx \frac{x_{n+1} - x_n}{\Delta t}
	\end{equation}
	
	However, this approach is inherently local, sensitive to high-frequency noise, and unstable over long sequences. It fails to capture the global structure of input variations, especially in the presence of correlated distractors or when long-term dynamics must be captured.

	\paragraph{Spectral Differentiation Principle.}
	To overcome these issues, we propose the \textbf{\textit{Spectral Differentiation Unit (SDU)}}, which estimates input derivatives using Fourier-based frequency-domain techniques. The method is grounded in the classical identity:
	
	\begin{equation}
	\mathcal{F} \left[ \frac{d}{dt} x(t) \right] = j\omega \cdot \mathcal{F}[x(t)]
	\end{equation}
	
	This identity implies that differentiation in the time domain corresponds to multiplication by $j\omega$ in the frequency domain. For discrete inputs, we apply the Discrete Fourier Transform (DFT), compute the frequency-wise product, and invert the result via the Inverse DFT (IDFT) to obtain an estimate of the derivative:
	
	\begin{equation}
	x'_n \approx \text{IDFT} \left( j \omega_k \cdot \text{DFT}(x_n) \right), \quad \omega_k = \frac{2\pi k}{N \Delta t}
	\end{equation}
	
	This method is known in numerical analysis as a pseudo-spectral method, and has the following advantages:
	
	\begin{itemize}
		\item \textbf{Global Receptive Field}: Fourier transforms incorporate the entire sequence, improving noise robustness.
		\item \textbf{Stable Estimation}: Avoids amplification of local perturbations common in finite differences.
		\item \textbf{Parallelizable via FFT}: Computationally efficient with complexity $O(N \log N)$.
		\item \textbf{Higher-Order Extensions}: Easily extended to second or higher-order derivatives via $(j\omega)^n$ scaling.
	\end{itemize}
	
	\paragraph{Application to Kalman-Optimal Dynamics.}
	In our model, the estimated derivative $\delta x_t$ is used to modulate the discrete-time state-space dynamics (from a zero-order hold discretization of Equation~\ref{eq:structured-ode}) as follows:
	
	\begin{equation}
	h_t = \overline{A}_K h_{t-1} + \overline{B}_K x_t + K \delta x_t,
	\label{eq:derivative Kalman-Optimal SSM}
    \end{equation}
	
	This formulation integrates spectral estimation directly into the state propagation, ensuring accurate temporal sensitivity. The detailed implementation of the SDU, including practical concerns such as frequency truncation and FFT stabilization, is presented in Appendix~\ref{appendix:sdu-details}, while 	Section~\ref{sec:frequency-response-of-sdu} (Figure~\ref{fig:sdu-response}). provides experimental validation confirming its effectiveness and stability.

	\subsection{Segment-wise Parallel Scan} \label{sec:segment pscan}
	While \textit{innovation}-dependent selectivity overcomes the limitations of input-only parameterization in models such as Mamba and Mamba-2, it introduces a new challenge: the dynamic coupling of the SSM parameters (e.g., \( A, B \)) with both the input content and the hidden state. This tight dependency disrupts the hardware-friendly assumptions underlying many efficient designs, such as fixed-kernel convolutions~\cite{krizhevsky2017imagenet,gu2022efficiently} and scan algorithms~\cite{gu2023mamba,dao2024transformers}, by preventing the precomputation of \( A_K \) and \( B_K \).
	
	To support efficient computation under the \textit{innovation}-modulated recurrence introduced in Section~\ref{sec:Improving SSMs}, we adopt a \textbf{\textit{Segment-wise Scan Algorithm}} balancing modeling expressiveness and computational efficiency. Specifically, we divide the input sequence \( X \in \mathbb{R}^{B \times L \times D} \) into \( M = \lceil L / S \rceil \) non-overlapping segments of length \( S \), applying:
	\begin{itemize}
		\item \textbf{Intra-segment}: parallel scan (vectorized over length \( S \)) to compute \( h_t \) and \( K \) simultaneously within each segment.
		\item \textbf{Inter-segment}: recursive updates across segments, using the final state of segment \(\ell\) as the initial condition of segment \(\ell+1\).
	\end{itemize}
	
	where \( \delta x_t \) is the input derivative estimated via SDU (Section~\ref{sec:SDU}), and the segment-wise scan is formalized as:
	\begin{equation}
		\textsc{Scan}(\overline{A}_K^{(\ell)}H^{(\ell-1)},~ \overline{B}_K^{(\ell)}X^{(\ell)} + K^{(\ell)} \delta X^{(\ell)}) \rightarrow H^{(\ell)}.
		\label{eq:ParallelScan}
	\end{equation}
	
	This approach reconciles Kalman-optimal propagation with modern hardware advantages by introducing a trade-off: longer segments improve parallelism at the cost of finer temporal granularity, while shorter segments increase sequential dependencies but preserve more precise state interactions.
	
	Table~\ref{tab:a} compares KOSS’s segment-wise scan with prior architectures, including full recurrence, global convolution, and hardware-aware scanning, highlighting how our approach balances parallel efficiency with dynamic context dependency. For detailed implementation strategies and discussions, see Appendix~\ref{appendix:segment-scan-details}.

		\begin{table}[!htbp]
		\centering
		\renewcommand{\arraystretch}{1.6}
		\setlength{\tabcolsep}{10pt}
		\begin{tabular}{@{}>{\centering\arraybackslash}p{3.8cm} 
				>{\centering\arraybackslash}p{1.6cm} 
				>{\centering\arraybackslash}p{2.2cm} 
				>{\centering\arraybackslash}p{2.2cm} 
				>{\centering\arraybackslash}p{3.2cm}@{}}
			\toprule
			\makecell{\textbf{Method}} 
			& \makecell{\textbf{Latency}\\\textbf{Depth}} 
			& \makecell{\textbf{Max}\\\textbf{Parallelism}} 
			& \makecell{\textbf{State}\\\textbf{Length}} 
			& \makecell{\textbf{Parameter}\\\textbf{Coupling}} \\
			\midrule
			
			\makecell{Full Recurrence}
			& \makecell{$L$} 
			& \makecell{$B$} 
			& \makecell{$L$} 
			& \makecell{$A, B$ fixed} \\
			
			\addlinespace
			\makecell{Global\\Convolution}
			& \makecell{$1$} 
			& \makecell{$B \times L$} 
			& \makecell{$L$} 
			& \makecell{$A, B$ fixed} \\
			
			\addlinespace
			\makecell{Hardware-aware}
			& \makecell{$\log L$}
			& \makecell{$B \times L$} 
			& \makecell{$\approx 1$} 
			& \makecell{$A, B \sim x_t$} \\
			
			\addlinespace
			\makecell{\textbf{Segment-wise}\\\textbf{Scan (Ours)}} 
			& \bm{$\frac{L}{S} + \log S}$
			& \bm{$B \times S$}
			& \bm{$\frac{L}{S}$}
			& \bm{$A, B, K \sim x_t, h_{t-1}$} \\
			
			\bottomrule
		\end{tabular}
		\caption{\textbf{Comparison with Prior Architectures:} 
			Comparison of model architectures in terms of latency, parallelism, hidden state requirements, and parameter coupling. Our method employs segment-wise prefix scans for intra-segment parallelism and Kalman-inspired recursion for inter-segment state propagation.
		}
		\label{tab:a}
	\end{table}

	\subsection{Layer Design} \label{sec:Layer Design}
\begin{figure}[!htbp]
		\centering
		\includegraphics[width=\textwidth]{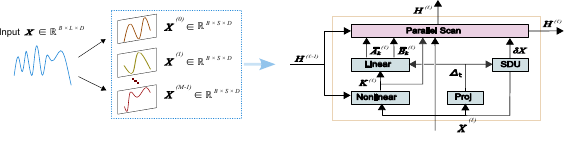}
		\caption{
			\textbf{Time-unrolled KOSS layer:} A nonlinear module estimates the Kalman gain \(\bm{K}^{(\ell)} \) from \(\bm{X}^{(\ell)}\)
			and \( \bm{H}^{(\ell-1)} \), which in turn modulates the dynamic parameters \( (\overline{\bm{A}}_K^{(\ell)}, \overline{\bm{B}}_K^{(\ell)}) \). The SDU computes input derivatives \( \delta \bm{X} \), which are used together with the modulated parameters in a parallel scan module to compute the updated hidden state \( \bm{H}^{(\ell)} \).}
		\label{fig:fig Layer Design}
	\end{figure}

	We construct a simple and homogeneous architecture by combining the Kalman-Optimal Selective mechanism with the MLP block of Transformers, forming a unified KOSS layer that integrates the core theoretical components introduced in Sections~\ref{sec:Kalman-Optimal Continuous SSMs} to~\ref{sec:segment pscan}, including Kalman-optimal information routing, Fourier-based input differentiation, and segment-wise parallel scanning. The resulting architecture demonstrates that Kalman-optimal design principles can be seamlessly integrated with classical selective state spaces for scalable long sequence modeling. See Figure~\ref{fig:fig Layer Design} for an overview of the time-unrolled KOSS layer.
	
	Each input segment \( \bm{X}^{(\ell)} \in \mathbb{R}^{B \times S \times D} \) is processed by the Spectral Differentiation Unit (SDU) to compute input derivatives \( \delta \bm{X} \)(~\ref{sec:SDU}), while a nonlinear module estimates the Kalman gain \( \bm{K}^{(\ell)} \) as a function of both input and hidden state \( \bm{H}^{(\ell-1)} \). The estimated gain \( \bm{K}^{(\ell)} \) determines the dynamic transition matrices \( (\overline{\bm{A}}_K^{(\ell)}, \overline{\bm{B}}_K^{(\ell)}) \) through the closed-form expressions derived in Equation~\ref{eq:structured-ode}. These parameters are passed to a parallel scan module that computes intra-segment state transitions and produces the updated recurrent state \( \bm{H}^{(\ell)}  \in \mathbb{R}^{B \times S \times D \times N} \)~(\ref{sec:segment pscan}).

\section{Empirical Evaluation} \label{sec:Empirical Evaluation}
	We conduct a comprehensive suite of experiments to evaluate the theoretical soundness, modeling capabilities, and practical effectiveness of KOSS. These experiments span multiple dimensions: 
	Section~\ref{sec:Synthetic Tasks} assesses the model’s ability to perform content- and context-aware selection under heavy distractor interference; 
	Section~\ref{sec:Main Experiments} presents large-scale forecasting experiments across six diverse domains, rigorously benchmarking KOSS against state-of-the-art baselines in terms of accuracy, stability, and cross-domain generalization;
	Section~\ref{sec:Segment Length Analysis} analyzes scalability by examining how varying segment lengths affect modeling fidelity and computational efficiency;
	Section~\ref{sec:Efficiency Benchmark} quantifies runtime and memory throughput to demonstrate hardware efficiency;
	Section~\ref{sec:Ablation Study} conducts ablation studies to isolate the roles of innovation-driven selectivity (IDS) and spectral differentiation (SDU); 
	finally, Sections~\ref{sec:convergence-of-Kalman-Gain} and~\ref{sec:frequency-response-of-sdu} empirically validate KOSS’s theoretical assumptions, including Kalman gain convergence and the frequency response of SDU. This multifaceted evaluation not only benchmarks performance, but also provides empirical validation of the theoretical properties established in ~\ref{sec:selective_mechanisms}, including optimal selective dynamics, robustness under disturbed sequences, and the spectral behavior induced by SDU.

	\subsection{Synthetic Tasks} \label{sec:Synthetic Tasks}
	The Copying task is a classic probe for memorization in sequence models. Prior work has shown that time-invariant SSMs like S4 can solve this task by tracking token positions rather than reasoning over content~\cite{gu2021efficiently,romero2021ckconv}. To prevent this shortcut, the \emph{Selective Copying} task randomizes token spacing, rendering simple timing cues ineffective. However, models like Mamba (with input-driven selectivity~\cite{dao2024transformers} can still solve it by directly identifying non-zero tokens, without understanding contextual relevance.
	
	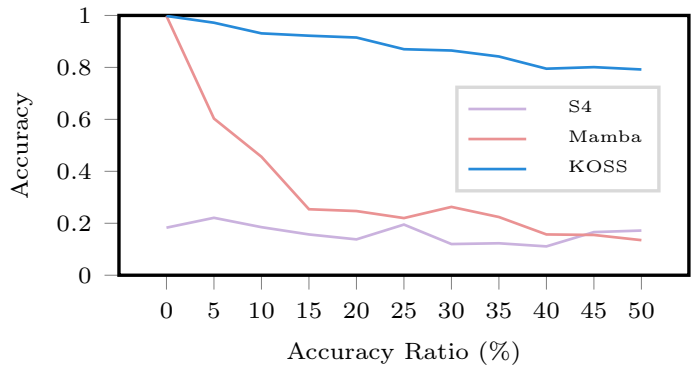
\begin{wrapfigure}{r}{0.6\textwidth}
		\vspace{-1.0em}
		\centering
		\definecolor{MyRed}{HTML}{EA9293}
		\definecolor{MyGreen}{HTML}{94CF95}
		\definecolor{MyPurple}{HTML}{CAB2DE}
		\resizebox{0.58\textwidth}{!}{%
			\begin{tikzpicture}
				\begin{axis}[
					width=7.5cm,
					height=4.3cm,
					xlabel={Accuracy Ratio (\%)},
					ylabel={Accuracy},
					xtick={0,5,10,15,20,25,30,35,40,45,50},
					ymin=0, ymax=1,
					axis lines=box,
					tick pos=left,
					xtick align=outside,
					ytick align=outside,
					xticklabel pos=left,
					yticklabel pos=bottom,
					legend style={
						at={(0.59,0.73)}, anchor=north west,
						font=\tiny,
						fill=white,
						draw=gray!30,
						legend cell align=left,
						column sep=1em,
						legend columns=1,
					},
					tick label style={font=\scriptsize},
					label style={font=\scriptsize},
					line width=1pt,
					mark size=1.5pt
					]
					\addplot[MyPurple, thick] coordinates {
						(0, 0.183) (5, 0.221) (10, 0.185) (15, 0.157) (20, 0.138)
						(25, 0.195) (30, 0.120) (35, 0.123) (40, 0.111) (45, 0.166) (50, 0.172)
					};
					\addlegendentry{S4}
					
					\addplot[MyRed, thick] coordinates {
						(0, 0.998) (5, 0.603) (10, 0.456) (15, 0.254) (20, 0.247)
						(25, 0.220) (30, 0.263) (35, 0.224) (40, 0.157) (45, 0.155)  (50, 0.135)
					}; 
					\addlegendentry{Mamba}
					
					\addplot[myblue, thick] coordinates {
						(0, 0.998) (5, 0.972) (10, 0.931) (15, 0.922) (20, 0.915)
						(25, 0.870) (30, 0.865) (35, 0.842) (40, 0.795) (45, 0.801) (50, 0.792)
					};
					\addlegendentry{KOSS}
				\end{axis}
			\end{tikzpicture}
		}
		\caption{\textbf{Selective Copying:} Performance under distractor interference. KOSS demonstrates strong resilience through innovation-driven selectivity, while Mamba and S4 degrade sharply under contextual noise. Interference is limited to 50\% to maintain a learnable signal-to-noise ratio.}
		\label{fig:Context-Aware Selective Copying Evaluation}
		\vspace{-1.0em}
	\end{wrapfigure}
	
	To evaluate deeper selection capability, we introduce \emph{correlated distractors}—non-zero tokens that mimic target patterns but carry no task-relevant information (Figure~\ref{fig:copy-vs-selective}). This invalidates naive heuristics and requires models to reason jointly over input and latent state.
	As shown in Figure~\ref{fig:Context-Aware Selective Copying Evaluation}, KOSS significantly outperforms both S4 and Mamba as interference increases, maintaining robust accuracy under heavy contextual noise (see Appendix~\ref{sec:Appendix Synthetic Tasks} for task setup and training protocols). For example, at 50\% interference—the hardest setting evaluated—KOSS retains 79.2\% accuracy, while Mamba and S4 drop to 13.5\% and 17.2\%, respectively. This result highlights KOSS’s ability to perform robust context-aware filtering under heavy distractor noise.

	\subsection{Main Experiment: Long-Term Forecasting} \label{sec:Main Experiments}
	To evaluate the ability of KOSS to model long-range dependencies and maintain stability over extended horizons, we conduct systematic long-term forecasting experiments on a broad set of real-world benchmarks.

	\paragraph{Datasets}
	We use nine widely adopted datasets spanning traffic, energy, economics, meteorology, and epidemiology. These datasets vary in sampling frequency, input dimensionality, and temporal structure, offering a comprehensive testbed for evaluating model robustness and generalization. Full dataset statistics and sources are provided in Appendix~\ref{sec:Appendix Main Experiments}.

	\paragraph{Baselines} 
	We compare KOSS against nine state-of-the-art (SOTA) models across four architectural families. To highlight differences in selective mechanisms, we first include two representative SSM-based approaches: FiLM~\cite{zhou2022film}, built on top of S4, and S-Mamba~\cite{wang2025mamba}, an extension of Mamba with input-driven selection. We further evaluate four Transformer-based models—FEDformer~\cite{zhou2022fedformer}, Autoformer~\cite{wu2021autoformer}, iTransformer~\cite{liu2023itransformer}, and Crossformer~\cite{zhang2023crossformer}—as well as two MLP-based methods DLinear~\cite{su2022dlinear}, PatchTST~\cite{huang2024long} and one TCN-based model TimeMixer~\cite{wang2024timemixer}. This selection ensures a comprehensive architectural comparison across the latest LTSF paradigms.
	
	\paragraph{Setup and Evaluation}
	We follow the experimental settings of S-Mamba~(\cite{wang2025mamba}) and tune the input sequence length for each task to optimize forecasting accuracy. By default, the prediction horizon is set to \( T \in \{96, 192, 336, 720\} \), with a segment length of \( S = 16 \). For the ILI dataset, which is sampled weekly, we use shorter prediction lengths \( T \in \{24, 36, 48, 60\} \) and a segment length \( S = 8 \).
	
	S-Mamba serves as the primary baseline due to its strong performance across most benchmarks. Complete implementation details and dataset statistics are provided in Appendix~\ref{sec:Appendix Main Experiments}.

	\renewcommand{\arraystretch}{1.22}
	\begin{table}[ht]
		\centering
		\caption{\textbf{Long-Term Forecasting:}
			Multivariate long-term forecasting results on the Traffic, Electricity, Exchange, Weather, and ILI datasets. 
			The prediction length is set to \( T \in \{96, 192, 336, 720\} \) for all datasets except ILI, which uses \( O \in \{24, 36, 48, 60\} \) due to its weekly resolution. 
			The best results are shown in \textbf{\textcolor{red}{bold red}}, and the second-best are \textcolor{MyPurple}{\underline{underlined purple}}. 
			All results are averaged over 5 runs.
		}
		
		\label{tab:traffic_results1}  % 可选：引用用的标签
		
		\scriptsize
		\resizebox{\textwidth}{!}{
			\begin{tabular}{c|c|cc|cc|cc|cc|cc|cc|cc|cc|cc|cc}
				\hline
				\multicolumn{2}{c|}{Models} & 
				\multicolumn{2}{c|}{\textbf{KOSS}} &
				\multicolumn{2}{c}{S-Mamba} &
				\multicolumn{2}{c|}{FiLM} & 
				\multicolumn{2}{c|}{iTransformer} &
				\multicolumn{2}{c|}{FEDFormer} &
				\multicolumn{2}{c|}{AutoFormer} &
				\multicolumn{2}{c|}{Crossformer} &
				\multicolumn{2}{c|}{DLinear} &
				\multicolumn{2}{c|}{PatchTST} &
				\multicolumn{2}{c}{TimeMixer} \\
				\hline
				
				\multicolumn{2}{c|}{Metric} & MSE & MAE & MSE & MAE & MSE & MAE & MSE & MAE & MSE & MAE & MSE & MAE & MSE & MAE & MSE & MAE & MSE & MAE & MSE & MAE \\
				\hline
				\multirow{4}{*}{\rotatebox{90}{Traffic}} 
				& 96  & \textcolor{red}{\textbf{0.288}} & \textcolor{red}{\textbf{0.226}} & \textcolor{MyPurple}{\underline{0.382}} & \textcolor{MyPurple}{\underline{0.261}}
				& 0.416 & 0.294 & 0.395 & 0.268 & 0.587 & 0.366 & 0.613 & 0.388 & 0.522 & 0.290 & 0.650 & 0.396 & 0.462 & 0.295 & 0.462 & 0.285 \\
				
				& 192 & \textcolor{red}{\textbf{0.250}} & \textcolor{red}{\textbf{0.210}} & \textcolor{MyPurple}{\underline{0.396}} & \textcolor{MyPurple}{\underline{0.267}} 
				& 0.408 & 0.288 & 0.417 & 0.276 & 0.604 & 0.373 & 0.616 & 0.382 & 0.530 & 0.293 & 0.598 & 0.370 & 0.466 & 0.296 & 0.473 & 0.296 \\
				
				& 336 & \textcolor{red}{\textbf{0.224}} & \textcolor{red}{\textbf{0.171}} & \textcolor{MyPurple}{\underline{0.417}} & \textcolor{MyPurple}{\underline{0.276}} 
				& 0.425 & 0.298 & 0.433 & 0.283 & 0.621 & 0.383 & 0.622 & 0.337 & 0.558 & 0.305 & 0.605 & 0.373 & 0.482 & 0.304 & 0.498 & 0.297 \\
				
				& 720 & \textcolor{red}{\textbf{0.292}} & \textcolor{red}{\textbf{0.230}} & \textcolor{MyPurple}{\underline{0.460}} & \textcolor{MyPurple}{\underline{0.300}} 
				& 0.520 & 0.353 & 0.467 & 0.302 & 0.626 & 0.382 & 0.660 & 0.408 & 0.589 & 0.328 & 0.645 & 0.394 & 0.514 & 0.322 & 0.506 & 0.313 \\
				\hline
				+36.23\% & Avg & \textcolor{red}{\textbf{0.264}} & \textcolor{red}{\textbf{0.209}} & \textcolor{MyPurple}{\underline{0.414}} & \textcolor{MyPurple}{\underline{0.276}} & 0.442 & 0.308 & 0.428 & 0.282 & 0.610 & 0.376 & 0.628 & 0.379 & 0.550 & 0.304 & 0.625 & 0.383 & 0.481 & 0.304 & 0.485 & 0.297 \\
				\hline

				\multirow{4}{*}{\rotatebox{90}{Electricity}} 
				& 96  & \textcolor{red}{\textbf{0.137}} & \textcolor{red}{\textbf{0.195}} & \textcolor{MyPurple}{\underline{0.139}} & \textcolor{MyPurple}{\underline{0.235}} & 0.154 & 0.267 & 0.148 & 0.240 & 0.193 & 0.308 & 0.201 & 0.317 & 0.219 & 0.314 & 0.197 & 0.282 & 0.181 & 0.270 & 0.153 & 0.247 \\
				
				& 192 & \textcolor{red}{\textbf{0.133}} & \textcolor{red}{\textbf{0.189}} & \textcolor{MyPurple}{\underline{0.159}} & 0.255 & 0.164 & 0.258 & 0.162 & \textcolor{MyPurple}{\underline{0.253}} & 0.201 & 0.315 & 0.222 & 0.334 & 0.231 & 0.322 & 0.196 & 0.285 & 0.188 & 0.274 & 0.166 & 0.256 \\
				
				& 336 & \textcolor{red}{\textbf{0.121}} & \textcolor{red}{\textbf{0.177}} & \textcolor{MyPurple}{\underline{0.176}} & 0.272 & 0.188 & 0.283 & 0.178 & \textcolor{MyPurple}{\underline{0.269}} & 0.214 & 0.329 & 0.231 & 0.338 & 0.246 & 0.337 & 0.209 & 0.301 & 0.204 & 0.293 & 0.185 & 0.277 \\
				
				& 720 & \textcolor{red}{\textbf{0.153}} & \textcolor{red}{\textbf{0.215}} & \textcolor{MyPurple}{\underline{0.204}} & \textcolor{MyPurple}{\underline{0.298}} & 0.236 & 0.332 & 0.225 & 0.317 & 0.246 & 0.355 & 0.254 & 0.361 & 0.280 & 0.363 & 0.245 & 0.333 & 0.246 & 0.324 & 0.225 & 0.310 \\
				\hline
				+20\% & Avg & \textcolor{red}{\textbf{0.136}} & \textcolor{red}{\textbf{0.194}} & \textcolor{MyPurple}{\underline{0.170}} & \textcolor{MyPurple}{\underline{0.265}} & 0.186 & 0.285 & 0.178 & 0.270 & 0.214 & 0.327 & 0.227 & 0.338 & 0.244 & 0.334 & 0.212 & 0.300 & 0.205 & 0.290 & 0.182 & 0.272 \\
				\hline

				\multirow{4}{*}{\rotatebox{90}{Exchange}} 
				& 96  & 0.121 & 0.217 & \textcolor{red}{\textbf{0.086}} & 0.207 & \textcolor{red}{\textbf{0.086}} & \textcolor{red}{\textbf{0.204}} & \textcolor{red}{\textbf{0.086}} & 0.206 & 0.148 & 0.278 & 0.197 & 0.323 & 0.256 & 0.367 & \textcolor{MyPurple}{\underline{0.088}} & 0.218 & \textcolor{MyPurple}{\underline{0.088}} & \textcolor{MyPurple}{\underline{0.205}} & 0.095 & 0.207 \\
				
				& 192 & 0.231 & \textcolor{red}{\textbf{0.291}} & 0.182 & 0.304 & 0.188 & \textcolor{MyPurple}{\textbf{0.292}} & 0.177 & 0.299 & 0.271 & 0.315 & 0.300 & 0.369 & 0.470 & 0.509 & \textcolor{MyPurple}{\underline{0.176}} & 0.315 & \textcolor{MyPurple}{\underline{0.176}} & 0.299 & \textcolor{red}{\textbf{0.151}} & 0.293 \\
				
				& 336 & 0.329 & \textcolor{red}{\textbf{0.340}} & 0.332 & 0.418 & 0.356 & 0.433 & 0.331 & 0.417 & 0.460 & 0.427 & 0.509 & 0.524 & 1.268 & 0.883 & 0.313 & 0.427 & \textcolor{MyPurple}{\underline{0.301}} & 0.397 & \textcolor{red}{\textbf{0.264}} & \textcolor{MyPurple}{\underline{0.361}} \\
				
				& 720 & \textcolor{red}{\textbf{0.383}} & \textcolor{red}{\textbf{0.408}} & 0.867 & 0.703 & 0.727 & 0.669 & 0.847 & 0.691 & 1.195 & 0.695 & 1.447 & 0.941 & 1.767 & 1.068 & 0.839 & 0.695 & 0.901 & 0.714 & \textcolor{MyPurple}{\underline{0.586}} & \textcolor{MyPurple}{\underline{0.602}} \\
				\hline
				+2.92\% & Avg & \textcolor{red}{\textbf{0.266}} & \textcolor{red}{\textbf{0.314}} & 0.367 & 0.408 & 0.339 & 0.400 & 0.360 & 0.403 & 0.519 & 0.429 & 0.613 & 0.539 & 0.940 & 0.707 & 0.354 & 0.414 & 0.367 & 0.404 & \textcolor{MyPurple}{\underline{0.274}} & \textcolor{MyPurple}{\underline{0.365}} \\
				\hline

				\multirow{4}{*}{\rotatebox{90}{Weather}} 
				& 96  & \textcolor{red}{\textbf{0.144}} & \textcolor{red}{\textbf{0.173}} & 0.165 & 0.210 & 0.199 & 0.262 & 0.174 & 0.214 & 0.217 & 0.296 & 0.266 & 0.336 & \textcolor{MyPurple}{\underline{0.158}} & 0.230 & 0.196 & 0.255 & 0.177 & 0.218 & 0.163 & \textcolor{MyPurple}{\underline{0.209}} \\
				
				& 192 & 0.216 & \textcolor{red}{\textbf{0.234}} & 0.214 & 0.252 & 0.228 & 0.288 & 0.221 & 0.254 & 0.276 & 0.336 & 0.307 & 0.367 & \textcolor{red}{\textbf{0.206}} & 0.277 & 0.237 & 0.296 & 0.225 & 0.259 & \textcolor{MyPurple}{\underline{0.208}} & \textcolor{MyPurple}{\underline{0.250}} \\
				
				& 336 & \textcolor{red}{\textbf{0.244}} & \textcolor{red}{\textbf{0.248}} & 0.274 & 0.297 & 0.267 & 0.323 & 0.278 & 0.296 & 0.339 & 0.380 & 0.359 & 0.395 & 0.272 & 0.335 & 0.283 & 0.335 & 0.278 & 0.297 & \textcolor{MyPurple}{\underline{0.251}} & \textcolor{MyPurple}{\underline{0.287}} \\
				
				& 720 & \textcolor{red}{\textbf{0.172}} & \textcolor{red}{\textbf{0.187}} & 0.350 & 0.345 & \textcolor{MyPurple}{\underline{0.319}} & 0.361 & 0.358 & 0.347 & 0.403 & 0.428 & 0.419 & 0.428 & 0.398 & 0.418 & 0.345 & 0.381 & 0.354 & 0.348 & 0.339 & \textcolor{MyPurple}{\underline{0.341}} \\
				\hline
				+19.17\% & Avg & \textcolor{red}{\textbf{0.194}} & \textcolor{red}{\textbf{0.211}} & 0.251 & 0.276 & 0.253 & 0.309 & 0.258 & 0.278 & 0.309 & 0.360 & 0.338 & 0.382 & 0.259 & 0.315 & 0.265 & 0.317 & 0.259 & 0.281 & \textcolor{MyPurple}{\underline{0.240}} & \textcolor{MyPurple}{\underline{0.271}} \\
				\hline

				\multirow{4}{*}{\rotatebox{90}{ILI}} 
				& 24  & \textcolor{red}{\textbf{1.158}} & \textcolor{red}{\textbf{0.506}} & -- & --  & 1.970 & 0.875 & 3.154 & 1.235 & 3.228 & 1.260 & 3.483 & 1.287 & 3.041 & 1.186 & 2.215 & 1.081 & \textcolor{MyPurple}{\underline{1.319}} & \textcolor{MyPurple}{\underline{0.754}} & 1.453 & 0.827 \\
				
				& 36 & \textcolor{red}{\textbf{1.271}} & \textcolor{red}{\textbf{0.564}} & -- & --  & 1.982 & \textcolor{MyPurple}{\underline{0.859}} & 2.544 & 1.083 & 2.679 & 1.150 & 3.103 & 1.148 & 3.406 & 1.232 & 1.963 & 0.963 & \textcolor{MyPurple}{\underline{1.579}} & 0.870 & 1.627 & 0.903 \\
				
				& 48 & \textcolor{red}{\textbf{1.108}} & \textcolor{red}{\textbf{0.545}} & -- & --  & 1.868 & 0.896 & 2.489 & 1.112 & 2.622 & 1.080 & 2.669 & 1.085 & 3.459 & 1.221 & 2.130 & 1.024 & \textcolor{MyPurple}{\underline{1.553}} & \textcolor{MyPurple}{\underline{0.815}} & 1.644 & 0.914 \\
				
				& 60 & \textcolor{red}{\textbf{1.119}} & \textcolor{red}{\textbf{0.557}} & -- & --  & 2.057 & 0.929 & 2.675 & 1.034 & 2.857 & 1.078 & 2.770 & 1.125 & 3.640 & 1.305 & 2.368 & 1.096 & \textcolor{MyPurple}{\underline{1.470}} & \textcolor{MyPurple}{\underline{0.788}} & 1.633 & 0.908 \\
				\hline
				+21.35\% & Avg & \textcolor{red}{\textbf{1.164}} & \textcolor{red}{\textbf{0.543}} & -- & --  & 1.969 & 0.890 & 2.715 & 1.116 & 2.847 & 1.170 & 3.006 & 1.161 & 3.387 & 1.236  & 2.169 & 1.041 & \textcolor{MyPurple}{\underline{1.480}} & \textcolor{MyPurple}{\underline{0.807}} & 1.589 & 0.888 \\
				\hline
				
				% 可以继续添加 Avg 等行
			\end{tabular}
		}
	\end{table}

	\begin{table}[ht]
		\centering
		\caption{\textbf{Long-Term Forecasting:}
			Forecasting results on the ETT datasets with prediction lengths \( T \in \{96, 192, 336, 720\} \). 
			The best results are shown in \textbf{\textcolor{red}{bold red}}, and the second-best are \textcolor{MyPurple}{\underline{underlined purple}}. 
			All results are averaged over 5 runs.
		}
		\label{tab:traffic_results2}
		
		\scriptsize
		\resizebox{\textwidth}{!}{
			\begin{tabular}{c|c|cc|cc|cc|cc|cc|cc|cc|cc|cc|cc}
				\hline
				\multicolumn{2}{c|}{Models} & 
				\multicolumn{2}{c|}{\textbf{KOSS}} &
				\multicolumn{2}{c}{S-Mamba} &
				\multicolumn{2}{c|}{FiLM} & 
				\multicolumn{2}{c|}{iTransformer} &
				\multicolumn{2}{c|}{FEDFormer} &
				\multicolumn{2}{c|}{AutoFormer} &
				\multicolumn{2}{c|}{Crossformer} &
				\multicolumn{2}{c|}{DLinear} &
				\multicolumn{2}{c|}{PatchTST} &
				\multicolumn{2}{c}{TimeMixer} \\
				\hline
				
				\multicolumn{2}{c|}{Metric} & MSE & MAE & MSE & MAE & MSE & MAE & MSE & MAE & MSE & MAE & MSE & MAE & MSE & MAE & MSE & MAE & MSE & MAE & MSE & MAE \\
				\hline
				
				\multirow{4}{*}{\rotatebox{90}{ETTm1}} 
				& 96  & \textcolor{red}{\textbf{0.215}} & \textcolor{red}{\textbf{0.224}} & 0.333 & 0.368 & -- & -- & 0.334 & 0.368 & 0.379 & 0.419 & 0.505 & 0.475 & 0.404 & 0.426 & 0.345 & 0.372 & 0.329 & 0.367 & \textcolor{MyPurple}{\underline{0.320}} & \textcolor{MyPurple}{\underline{0.357}} \\
				
				& 192 & \textcolor{red}{\textbf{0.230}} & \textcolor{red}{\textbf{0.237}} & 0.376 & 0.390 & -- & -- & 0.377 & 0.391 & 0.426 & 0.441 & 0.553 & 0.496 & 0.450 & 0.451 & 0.380 & 0.389 & 0.367 & 0.385 & \textcolor{MyPurple}{\underline{0.361}} & \textcolor{MyPurple}{\underline{0.381}} \\
				
				& 336 & 0.424 & \textcolor{red}{\textbf{0.339}} & 0.408 & 0.413 & -- & -- & 0.426 & 0.420 & 0.445 & 0.459 & 0.621 & 0.537 & 0.532 & 0.515 & 0.413 & 0.413 & \textcolor{MyPurple}{\underline{0.399}} & 0.410 & \textcolor{red}{\textbf{0.390}} & \textcolor{MyPurple}{\underline{0.404}} \\
				
				& 720 & 0.490 & \textcolor{red}{\textbf{0.365}} & 0.475 & 0.448 & -- & -- & 0.491 & 0.459 & 0.543 & 0.490 & 0.671 & 0.561 & 0.666 & 0.589 & 0.474 & 0.453 & \textcolor{red}{\textbf{0.454}} & \textcolor{MyPurple}{\underline{0.439}} & \textcolor{MyPurple}{\underline{0.458}} & 0.441 \\
				\hline
				+10.99\% 
				& Avg & \textcolor{red}{\textbf{0.340}} & \textcolor{red}{\textbf{0.291}} & 0.398 & 0.405 & -- & -- & 0.407 & 0.410 & 0.448 & 0.452 & 0.588 & 0.517 & 0.513 & 0.496 & 0.403 & 0.407 & 0.387 & 0.400 & \textcolor{MyPurple}{\underline{0.382}} & \textcolor{MyPurple}{\underline{0.395}} \\
				\hline

				\multirow{4}{*}{\rotatebox{90}{ETTm2}} 
				& 96  & \textcolor{red}{\textbf{0.110}} & \textcolor{red}{\textbf{0.234}} & 0.179 & 0.263 & \textcolor{MyPurple}{\underline{0.165}} & \textcolor{MyPurple}{\underline{0.256}} & 0.180 & 0.264 & 0.203 & 0.287 & 0.255 & 0.339 & 0.287 & 0.366 & 0.193 & 0.292 & 0.175 & 0.259 & 0.175 & 0.258 \\
				
				& 192 & \textcolor{red}{\textbf{0.132}} & \textcolor{red}{\textbf{0.273}} & 0.250 & 0.309 & \textcolor{MyPurple}{\underline{0.222}} & \textcolor{MyPurple}{\underline{0.296}} & 0.250 & 0.309 & 0.269 & 0.328 & 0.281 & 0.340 & 0.414 & 0.492 & 0.284 & 0.362 & 0.241 & 0.302 & 0.237 & 0.299 \\
				
				& 336 & \textcolor{red}{\textbf{0.234}} & 0.382 & 0.312 & 0.349 & \textcolor{MyPurple}{\underline{0.277}} & \textcolor{red}{\textbf{0.333}} & 0.311 & 0.348 & 0.325 & 0.366 & 0.339 & 0.372 & 0.597 & 0.542 & 0.369 & 0.427 & 0.305 & 0.343 & 0.298 & \textcolor{MyPurple}{\underline{0.340}} \\
				
				& 720 & \textcolor{red}{\textbf{0.258}} & 0.419 & 0.411 & 0.406 & 0.371 & \textcolor{MyPurple}{\underline{0.389}} & 0.412 & 0.407 & 0.421 & 0.415 & 0.433 & 0.432 & 1.730 & 1.042 & 0.554 & 0.522 & 0.402 & 0.400 & \textcolor{MyPurple}{\underline{0.275}} & \textcolor{red}{\textbf{0.323}} \\
				\hline
				+25.20\% 
				& Avg & \textcolor{red}{\textbf{0.184}} & 0.327 & 0.288 & 0.332 & 0.259 & \textcolor{MyPurple}{\underline{0.319}} & 0.288 & 0.332 & 0.305 & 0.349 & 0.327 & 0.371 & 0.757 & 0.610 & 0.350 & 0.401 & 0.281 & 0.326 & \textcolor{MyPurple}{\underline{0.246}} & \textcolor{red}{\textbf{0.306}} \\
				\hline

				\multirow{4}{*}{\rotatebox{90}{ETTh1}} 
				& 96  & \textcolor{red}{\textbf{0.298}} & \textcolor{red}{\textbf{0.267}} & 0.386 & 0.405 & -- & -- & 0.386 & 0.405 & 0.376 & 0.419 & 0.449 & 0.459 & 0.423 & 0.448 & 0.386 & \textcolor{MyPurple}{\underline{0.400}} & 0.414 & 0.419 & \textcolor{MyPurple}{\underline{0.375}} & 0.400 \\
				
				& 192 & \textcolor{red}{\textbf{0.331}} & \textcolor{red}{\textbf{0.304}} & 0.443 & 0.437 & -- & -- & 0.441 & 0.436 & 0.420 & 0.448 & 0.500 & 0.482 & 0.471 & 0.474 & \textcolor{MyPurple}{\underline{0.437}} & 0.432 & 0.460 & 0.445 & 0.479 & \textcolor{MyPurple}{\underline{0.421}} \\
				
				& 336 & \textcolor{red}{\textbf{0.388}} & \textcolor{red}{\textbf{0.323}} & 0.489 & 0.468 & -- & -- & 0.487 & \textcolor{MyPurple}{\underline{0.458}} & \textcolor{MyPurple}{\underline{0.459}} & 0.465 & 0.521 & 0.496 & 0.570 & 0.546 & 0.481 & 0.459 & 0.501 & 0.466 & 0.484 & \textcolor{MyPurple}{\underline{0.458}} \\
				
				& 720 & \textcolor{red}{\textbf{0.471}} & \textcolor{red}{\textbf{0.368}} & 0.502 & 0.489 & -- & -- & 0.503 & 0.491 & 0.506 & 0.507 & 0.514 & 0.512 & 0.653 & 0.621 & 0.519 & 0.516 & 0.500 & 0.488 & \textcolor{MyPurple}{\underline{0.498}} & \textcolor{MyPurple}{\underline{0.482}} \\
				\hline
				+15.45\% 
				& Avg & \textcolor{red}{\textbf{0.372}} & \textcolor{red}{\textbf{0.316}} & 0.455 & 0.450 & -- & -- & 0.454 & 0.447 & \textcolor{MyPurple}{\underline{0.440}} & 0.460 & 0.496 & 0.487 & 0.529 & 0.522 & 0.456 & 0.452 & 0.469 & 0.454 & 0.459 & \textcolor{MyPurple}{\underline{0.440}} \\
				\hline

				\multirow{4}{*}{\rotatebox{90}{ETTh2}} 
				& 96  & \textcolor{red}{\textbf{0.192}} & \textcolor{red}{\textbf{0.338}} & 0.296 & 0.348 & -- & -- & 0.297 & 0.349 & 0.358 & 0.397 & 0.346 & 0.388 & 0.745 & 0.584 & 0.333 & 0.387 & 0.302 & 0.348 & \textcolor{MyPurple}{\underline{0.289}} & \textcolor{MyPurple}{\underline{0.341}} \\
				
				& 192 & \textcolor{red}{\textbf{0.236}} & \textcolor{red}{\textbf{0.385}} & 0.376 & 0.396 & -- & -- & 0.380 & 0.400 & 0.429 & 0.439 & 0.456 & 0.452 & 0.877 & 0.656 & 0.477 & 0.476 & 0.388 & 0.400 & \textcolor{MyPurple}{\underline{0.372}} & \textcolor{MyPurple}{\underline{0.392}} \\
				
				& 336 & \textcolor{red}{\textbf{0.258}} & \textcolor{red}{\textbf{0.393}} & 0.424 & 0.431 & -- & -- & 0.428 & 0.432 & 0.496 & 0.487 & 0.482 & 0.486 & 1.043 & 0.731 & 0.594 & 0.541 & 0.426 & 0.433 & \textcolor{MyPurple}{\underline{0.386}} & \textcolor{MyPurple}{\underline{0.414}} \\
				
				& 720 & \textcolor{red}{\textbf{0.314}} & \textcolor{red}{\textbf{0.423}} & 0.426 & 0.444 & -- & -- & 0.427 & 0.445 & 0.463 & 0.474 & 0.515 & 0.511 & 1.104 & 0.763 & 0.831 & 0.657 & 0.431 & 0.446 & \textcolor{MyPurple}{\underline{0.412}} & \textcolor{MyPurple}{\underline{0.434}} \\
				\hline
				+31.31\% 
				& Avg & \textcolor{red}{\textbf{0.250}} & \textcolor{red}{\textbf{0.385}} & 0.381 & 0.405 & -- & -- & 0.383 & 0.407 & 0.437 & 0.449 & 0.450 & 0.459 & 0.942 & 0.684 & 0.559 & 0.515 & 0.387 & 0.407 & \textcolor{MyPurple}{\underline{0.364}} & \textcolor{MyPurple}{\underline{0.395}} \\
				\hline
				
				% 可以继续添加 Avg 等行
			\end{tabular}
		}
	\end{table}

	\subsubsection{Overall Performance} \label{sec:Overall Performance}
	Table~\ref{tab:traffic_results1} and Table~\ref{tab:traffic_results2} present a comparative analysis of the overall performance of our models and various baselines across all datasets. We adopt two widely used metrics, MSE and MAE, to evaluate the model’s performance.
	
	From the reported results, we summarize three observations and attach the analysis: 
	\begin{itemize}
		\item \textbf{KOSS consistently outperforms all baseline models across nine real-world datasets.} In domains such as traffic (+36.23\%), electricity (+20\%), weather (+19.17\%), public health (+21.35\%), and industrial energy consumption—ETTm1 (+10.99\%), ETTm2 (+25.20\%), ETTh1 (+15.45\%), and ETTh2 (+31.31\%)—KOSS achieves over 20\% average improvement. 
		These datasets span diverse statistical characteristics, including high dimensionality, mixed periodicity, and varying noise levels. 
		The Kalman-inspired architecture of KOSS efficiently captures such structured patterns via rapid Kalman gain convergence and robust state updates, demonstrating strong cross-domain generalization and resilience.

		\item \textbf{Unlike most baselines, KOSS maintains or improves performance as the prediction horizon increases.}  
		While longer horizons typically degrade accuracy in other models, KOSS benefits from the convergence of the Kalman gain \( K(t) \), as analyzed in Section~\ref{sec:Kalman-Optimal Continuous SSMs}.  
		These dynamics indicate that the model steadily adapts to the input distribution and noise over time, which facilitates stable long-range dependency modeling and accurate long-term forecasts; see Appendix~\ref{appendix:kalman-derivative} for a detailed justification of this convergence property and steady-state approximation.
		
		\item \textbf{KOSS yields a significantly narrower gap between MSE and MAE compared to other models,} 
		suggesting both lower variance and better control of extreme prediction errors. This indicates that KOSS achieves not only smaller average errors but also greater robustness to outliers and noise. Such error balance is critical for long-term forecasting, where deviations may compound over time. The Kalman-inspired selective dynamics are central to controlling both error magnitude and variance.
		
	\end{itemize}
	
\begin{figure*}[t]
			\centering
			
			% ---------- Row 1: KOSS ----------
			\begin{minipage}{0.19\textwidth}
				\centering
				\includegraphics[width=\linewidth]{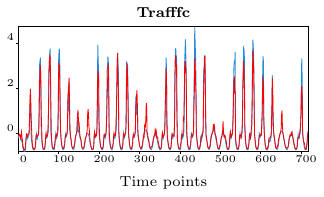}
			\end{minipage}
			\hfill
			\begin{minipage}{0.19\textwidth}
				\centering
				\includegraphics[width=\linewidth]{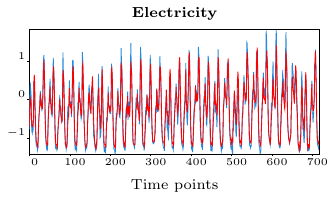}
			\end{minipage}
			\hfill
			\begin{minipage}{0.19\textwidth}
				\centering
				\includegraphics[width=\linewidth]{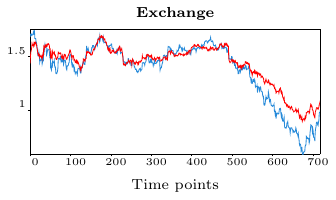}
			\end{minipage}
			\hfill
			\begin{minipage}{0.19\textwidth}
				\centering
				\includegraphics[width=\linewidth]{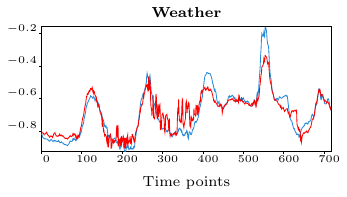}
			\end{minipage}
			\hfill
			\begin{minipage}{0.19\textwidth}
				\centering
				\includegraphics[width=\linewidth]{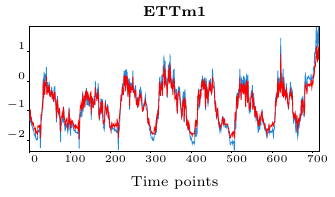}
			\end{minipage}
			{\small (a) KOSS}\\[0.5em]
			\vspace{0.8em}
			
			% ---------- Row 2: S-Mamba ----------
			\begin{minipage}{0.19\textwidth}
				\centering
				\includegraphics[width=\linewidth]{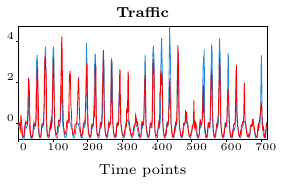}
			\end{minipage}
			\hfill
			\begin{minipage}{0.19\textwidth}
				\centering
				\includegraphics[width=\linewidth]{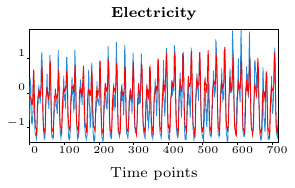}
			\end{minipage}
			\hfill
			\begin{minipage}{0.19\textwidth}
				\centering
				\includegraphics[width=\linewidth]{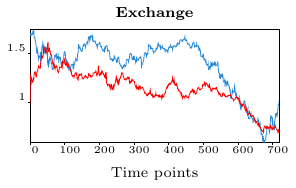}
			\end{minipage}
			\hfill
			\begin{minipage}{0.19\textwidth}
				\centering
				\includegraphics[width=\linewidth]{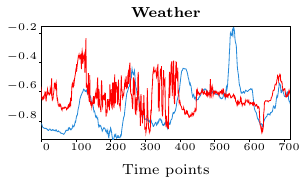}
			\end{minipage}
			\hfill
			\begin{minipage}{0.19\textwidth}
				\centering
				\includegraphics[width=\linewidth]{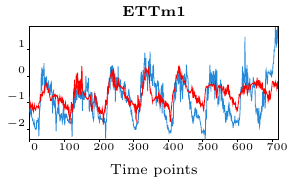}
			\end{minipage}
			\hspace{1.5em}{\small (b) S-Mamba}\\[0.5em]
			
			\caption{\textbf{KOSLM vs. S-Mamba:}
				Forecasting comparison on five representative datasets with both input and prediction horizons set to 720.
				\textbf{\textcolor{myblue}{the blue line}} denotes the ground truth, and \textbf{\textcolor{red}{the red line}} indicates model predictions.
				KOSS exhibits superior long-horizon stability and trend consistency compared to S-Mamba.}
			\label{fig:720_forecast_results}
			
		\end{figure*}

	To complement the quantitative results reported in Tables~\ref{tab:traffic_results1} and~\ref{tab:traffic_results2}, we further provide qualitative visualizations of long-horizon forecasting trajectories at the maximum prediction length ($T=720$).
	Figure~\ref{fig:720_forecast_results} illustrates representative forecasting cases on five datasets, comparing KOSS with S-Mamba.
	
	As shown in the figure, KOSS produces predictions that remain closely aligned with the ground truth in both trend and amplitude over extended horizons.
	In particular, for relatively smooth and weakly non-stationary series such as \textit{Weather} and \textit{Exchange}, KOSS exhibits reduced temporal lag and mitigated amplitude decay. By contrast, S-Mamba shows noticeable attenuation and phase shift as the forecasting horizon increases. These visual patterns are consistent with the quantitative performance trends observed in the main experiments, and provide an intuitive illustration of KOSS’s improved long-horizon forecasting behavior.
	 
	\subsection{Ablation Study} \label{sec:Ablation Study}
	To assess the contributions of key architectural components in KOSS, we conduct ablation studies on two core modules: the Innovation-Driven Selectivity (IDS) mechanism (Section~\ref{sec:Improving SSMs}) and the Spectral Differentiation Unit (SDU) (Section~\ref{sec:SDU}).
	
	\paragraph{Ablation of IDS} We replace the IDS module with several alternatives, including MLP, Attention, LSTM, S4, and S6. MLP serves as a neutral baseline that removes all structural inductive bias and selectivity. Attention is included for its strong capacity to extract global temporal patterns. LSTM offers recurrent modeling with implicit memory, while S4 and S6 represent structure-aware and input-dependent SSMs that offer different forms of dynamic selection. These variants are used to evaluate the role of context-aware selectivity in driving model performance. As shown in Table~\ref{tab:Ablation_IDS}, replacing IDS with any of the alternatives leads to performance degradation in both MSE and MAE, particularly on longer horizons. This confirms the critical role of innovation-driven selectivity in enabling precise long-term forecasting.

	\begin{table}[ht]
		\centering
		\caption{(\textbf{Ablations: IDS.})
			IDS is replaced with MLP, Attention, S4, and S6 on the Weather and Exchange datasets. The table reports MSE and MAE as relative differences from the original KOSS model (\( + \) indicates degraded performance, \( - \) indicates improved performance). All experiments use forecasting horizons \( T \in \{96, 192, 336, 720\} \).
		}
		\label{tab:Ablation_IDS}
		\scriptsize
		\resizebox{\textwidth}{!}{
			\begin{tabular}{c|c|cc|cc|cc|cc|cc}
				\hline
				\multicolumn{2}{c|}{Methods} & 
				\multicolumn{2}{c|}{\textbf{KOSS}} &
				\multicolumn{2}{c|}{MLP} &
				\multicolumn{2}{c|}{Attention} & 
				\multicolumn{2}{c|}{S4} &
				\multicolumn{2}{c}{S6} \\
				\hline
				
				\multicolumn{2}{c|}{Metric} & MSE & MAE & MSE & MAE & MSE & MAE & MSE & MAE & MSE & MAE \\
				\hline
				
				\multirow{4}{*}{{Weather}} 
				& 96  & 0.144 & 0.171 & +6.37\% & +6.67\% & -2.78\% & -3.39\% & -0.99\% & -1.70\% & -1.14\% & -2.14\% \\
				
				& 192 & 0.224 & 0.236 & +25.44\% & +15.98\% & -22.58\% & -11.13\% & +16.93\% & +15.82\% & +19.31\% & +4.24\% \\
				
				& 336 & 0.249 & 0.248 & +25.75\% & +13.69\% & -31.39\% & -14.91\% & +28.85\% & +14.96\% & +30.35 \% & +9.53\% \\
				
				& 720 & 0.169 & 0.175 & +37.78\% & +44.67\% & -9.37\% & -6.03\% & +1.024\% & +1.71\% & +1.18\% & +1.50\% \\
				\hline

				\multirow{4}{*}{{Exchange}} 
				& 96  & 0.135 & 0.212 & +31.85\% & +14.03\% & +55.14\% & +24.26\% & +75.36\% & +27.60\% & +33.53\% & +13.90\% \\
				
				& 192 & 0.279 & 0.292 & +43.93\% & +26.09\% & +19.99\% & +14.26\% & +12.26\% & +12.85\% & +19.58\% & +16.96\% \\
				
				& 336 & 0.341 & 0.339 & +51.38\% & +23.80\% & +47.42\% & +22.66\% & +56.56\% & +27.81\% & +87.63\% & +38.11\% \\
				
				& 720 & 0.282 & 0.322 & +12.69\% & -0.58\% & -2.03\% & -4.19\% & +29.93\% & +10.04\% & +18.05\% & +3.72\% \\
				\hline
				
				% 可以继续添加 Avg 等行
			\end{tabular}
		}
	\end{table}

	\begin{table}[ht]
		\centering
		\caption{(\textbf{Ablations: SDU.}) 
			SDU is replaced with MLP, LSTM, and Attention, or removed entirely (w/o), on the Weather and Exchange datasets. The table reports MSE and MAE as relative differences from the original KOSS model (\( + \) indicates degraded performance, \( - \) indicates improved performance). All experiments use forecasting horizons \( T \in \{96, 192, 336, 720\} \).}
		\label{tab:Ablation_SDU}
		
		\scriptsize
		\resizebox{\textwidth}{!}{
			\begin{tabular}{c|c|cc|cc|cc|cc|cc}
				\hline
				\multicolumn{2}{c|}{Methods} & 
				\multicolumn{2}{c|}{\textbf{KOSS}} &
				\multicolumn{2}{c|}{MLP} & 
				\multicolumn{2}{c|}{Attention} &
				\multicolumn{2}{c}{LSTM} &
				\multicolumn{2}{c|}{w/o}  \\
				\hline
				
				\multicolumn{2}{c|}{Metric} & MSE & MAE & MSE & MAE & MSE & MAE & MSE & MAE & MSE & MAE \\
				\hline
				
				\multirow{4}{*}{{Weather}} 
				& 96  &0.144 & 0.171 & +6.26\% & +5.87\% & +0.77\% & +0.71\% & +0.75\% & +0.89\% & -3.25\% & -1.97\% \\
				
				& 192 & 0.224 & 0.236 & +4.22\% & +0.14\% & +2.54\% & -0.68\% & +15.27\% & +4.35\% & +11.91\% & +2.25\% \\
				
				& 336 & 0.249 & 0.248 & +7.08\% & +5.77\% & +28.00\% & +10.91\% & +18.80\% & +10.43\% & +25.87\% & +9.90\% \\
				
				& 720 & 0.169 & 0.175 & +18.50\% & -1.18\% & -1.14\% & -0.60\% & +0.0\% & +0.62\% & +1.69\% & +2.74\% \\
				\hline

				\multirow{4}{*}{{Exchange}} 
				& 96  & 0.135 & 0.212 & +9.58\% & +5.55\% & +30.88\% & +17.14\% & -0.13\% & -2.35\% & +30.70\% & +17.10\% \\
				
				& 192 & 0.279 & 0.292 & +19.75\% & +18.91\% & +22.17\% & +7.39\% & -7.49\% & -0.99\% & +22.95\% & +19.84\% \\
				
				& 336 & 0.341 & 0.339 & +15.35\% & +9.60\% & +12.47\% & +8.90\% & +34.83\% & +17.26\% & +26.39\% & +15.74\% \\
				
				& 720 & 0.282 & 0.322 & +3.50\% & +2.29\% & -10.49\% & -4.89\% & +2.87\% & -2.51\% & +39.86\% & +16.66\% \\
				\hline
				
				% 可以继续添加 Avg 等行
			\end{tabular}
		}
	\end{table}
	
	\paragraph{Ablation of SDU} 
	We conduct ablation studies to assess both the effectiveness and necessity of the Fourier Differentiation Unit (SDU) in our Kalman-inspired framework.  
	\begin{itemize}
		\item \textbf{To evaluate the effectiveness of derivative estimation,} we replace SDU with generic deep networks such as MLP, LSTM, and Attention, which are widely used and possess strong sequence modeling capabilities. This setup provides a strong baseline for evaluating whether  SDU can deliver more accurate and stable derivative estimates than widely used deep sequence models with high expressive capacity.
		\item \textbf{To assess the necessity of derivative modeling itself,} we remove SDU entirely, eliminating the derivative term from the Kalman-inspired update equation.
	\end{itemize}
	Table~\ref{tab:Ablation_SDU} shows consistent performance drops for all replacements, particularly over longer horizons, suggesting that generic deep models cannot match the precision of Fourier-based estimation. The most significant decline occurs when SDU is removed, highlighting the structural importance of input derivative modeling in our framework.

	 \subsection{Efficiency Benchmark}  \label{sec:Efficiency Benchmark}
	 \paragraph{Runtime Benchmarks} 
	 We benchmark the speed of the SSM scan operation on the ETTm2 dataset using a batch size of 32 and input sequences of length 1024, processed with a segment length of \( S = 32 \), as shown in Figure~\ref{fig:scan-vs-conv}. 
	 Our efficient segment-wise SSM scan achieves a 5–20× speedup over the standard scan implementation in PyTorch.
	 While it is 5–10× slower than Mamba’s hardware-aware kernel, it  exhibits linear scaling comparable to global convolution-based methods.

	 \paragraph{Memory Benchmarks}
	 In addition, we report GPU memory footprint under standard training settings~\cite{zhang2023xformer} as a complementary baseline for efficiency analysis, as shown in Figure~\ref{fig:Memory benchmark}. For KOSS, we include two variants with different segment lengths (\( S = 16, 32 \)) to examine memory scalability under varying levels of granularity.

	 \begin{figure}[ht]
	 	\centering
	 	\begin{minipage}[t]{0.48\linewidth}
	 		\centering
	 		\begin{tikzpicture}[scale=0.5]
	 			\begin{loglogaxis}[
	 				width=12cm,
	 				height=7cm,
	 				xlabel={Sequence length},
	 				ylabel={Time (ms)},
	 				title={Scan vs Convolution},
	 				legend style={
	 					at={(0.03,0.97)}, 
	 					draw=gray!30, % 设置边框颜色为淡灰色
	 					anchor=north west, 
	 					font=\small},
	 				xtick={512,1000,2000,4000,8000,16000,32000,64000,128000},
	 				xticklabels={512,1k,2k,4k,8k,16k,32k,64k,128k},
	 				ytick={0.1, 1, 10, 100, 1000},
	 				yticklabels={0.1, 1, 10, 100, 1000},
	 				ymin=0.1, ymax=2000,
	 				log ticks with fixed point,
	 				grid=none,
	 				axis lines=box,
	 				tick pos=left,
	 				xtick align=outside,
	 				ytick align=outside,
	 				xticklabel pos=left,
	 				yticklabel pos=bottom,
	 				]
	 				\addlegendentry{Convolution}
	 				\addplot[orange, thick] coordinates {
	 					(512, 0.612) (1000, 0.78) (2000, 1.62) (4000, 4.68)
	 					(8000, 9.42) (16000, 23.4) (32000, 53.82) (64000, 105.6)
	 				};
	 				\addlegendentry{Scan (PyTorch)}
	 				\addplot[mygreen, thick] coordinates {
	 					(512, 6.18) (1000, 11.64) (2000, 23.7) (4000, 54.6)
	 					(8000, 120.0) (16000, 258.0) (32000, 550.2)
	 				};
	 				\addlegendentry{Scan (Mamba)}
	 				\addplot[red, thick] coordinates {
	 					(512, 0.15) (1000, 0.258) (2000, 0.45) (4000, 1.02)
	 					(8000, 2.22) (16000, 5.706) (32000, 12.48) (64000, 25.68)
	 				};
	 				\addlegendentry{Scan (Ours)}
	 				\addplot[myblue, thick] coordinates {
	 					(512, 1.68) (1000, 2.01348) (2000, 2.65752) (4000, 3.936)
	 					(8000, 12.8034) (16000, 44.013) (32000, 114.9) (64000, 243.9)
	 				};
	 			\end{loglogaxis}
	 		\end{tikzpicture}
	 		\caption{(\textbf{Runtime Benchmarks.}) Our efficient scan is up to 20× faster than a standard PyTorch implementation during training.}
	 		\label{fig:scan-vs-conv}
	 	\end{minipage}
	 	\hfill
	 	\begin{minipage}[t]{0.48\linewidth}
	 		\centering
	 		\begin{tikzpicture}[scale=0.5]
	 			\begin{axis}[
	 				width=15cm,
	 				height=7cm,
	 				title={Memory Footprint (Input Length = 1024)},
	 				xlabel={Batch size},
	 				ylabel={Memory Usage (GB)},
	 				ymin=0, ymax=7,
	 				enlarge x limits=0.1,
	 				bar width=7pt,
	 				ybar,
	 				symbolic x coords={1,2,4,8,16,32,64,128},
	 				xtick=data,
	 				tick label style={font=\small},
	 				nodes near coords={\pgfmathprintnumber\pgfplotspointmeta},
	 				nodes near coords align={vertical},
	 				nodes near coords style={
	 					rotate=90,
	 					anchor=west,
	 					font=\tiny,
	 					text=black  % 添加此行以统一为黑色字体
	 				},
	 				/pgf/number format/.cd,
	 				fixed,
	 				precision=2,
	 				legend style={
	 					at={(0.02,0.98)},
	 					anchor=north west,
	 					font=\small,
	 					draw=gray!30, % 设置边框颜色为淡灰色
	 					fill=white    % 背景填充为白色（可选）
	 				},
	 				axis lines=box,
	 				xtick pos=bottom,
	 				ytick pos=left,
	 				tick align=outside,
	 				]

	 				% Mamba (MB / 1024)
	 				\addplot+[fill=myblue, draw=none] coordinates {
	 					(1,0.107) (2,0.213) (4,0.401) (8,0.788) (16,1.55) (32,3.08) (64,6.14)
	 				};
	 				
	 				% Transformer
	 				\addplot+[fill=orange, draw=none] coordinates {
	 					(1,0.134) (2,0.152) (4,0.183) (8,0.307) (16,0.572) (32,1.11) (64,2.2)
	 				};

	 				% KOSS-16
	 				\addplot+[fill=mygreen, draw=none] coordinates {
	 					(1,0.033) (2,0.0498) (4,0.083) (8,0.148) (16,0.384) (32,0.542) (64,1.099)
	 				};
	 				
	 				% KOSS-32
	 				\addplot+[fill=red, draw=none] coordinates {
	 					(1,0.034) (2,0.052) (4,0.087) (8,0.158) (16,0.278) (32,0.566) (64,1.137)
	 				};

	 				\legend{Mamba, Transformer, KOSS-16, KOSS-32}
	 			\end{axis}
	 		\end{tikzpicture}
	 		
	 		\caption{(\textbf{Memory Benchmarks.}) Training: KOSS can achieves up to 6× lower GPU memory usage than Mamba, despite both being recurrent models.}
	 		\label{fig:Memory benchmark}
	 	\end{minipage}
	 \end{figure}
	 
	 \begin{table}[ht]
	 	\centering
	 	\caption{Parameter sizes of KOSS and baseline models.}
	 	\label{tab:parameter size}
	 	\begin{tabular}{c|ccccccccc}
	 		\toprule
	 		Methods & Transformer & Mamba & FiLM & KOSS \\
	 		\midrule
	 		Parameter & 1.17M & 0.07M & 1.50M & 0.2M  \\
	 		\bottomrule
	 	\end{tabular}
	 \end{table}

	 Notably, KOSS maintains a relatively small number of trainable parameters (\textbf{0.2M}, compared to Transformer’s 1.17M and Mamba’s 0.07M) and simultaneously achieves the lowest memory footprint among all evaluated models, as summarized in Table~\ref{tab:parameter size} and Figure~\ref{fig:Memory benchmark}. \textbf{\textit{This highlights a key advantage of our design: KOSS demonstrates strong performance while remaining lightweight in both parameter count and memory consumption.}} Unlike baseline models that rely on parameter-heavy or memory-intensive architectures, KOSS employs a structured, segment-wise formulation that enables dynamic computation with minimal overhead. This efficient architecture is well-suited for long-sequence modeling tasks, providing a favorable balance between scalability and generalization.

	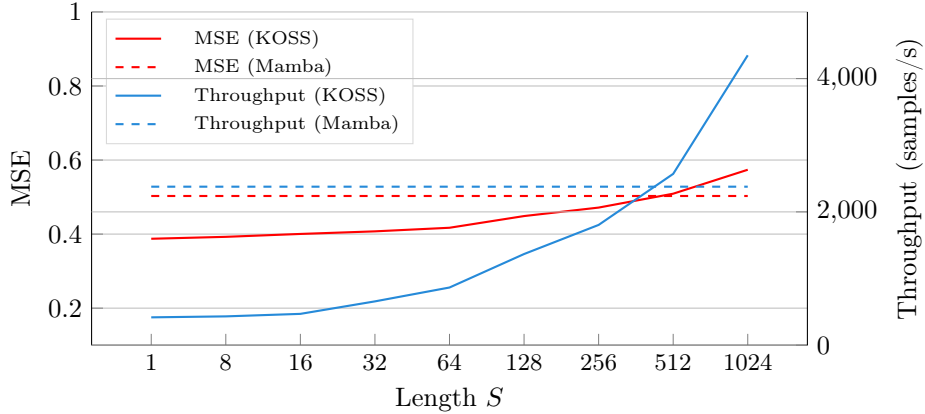
\begin{figure}[ht]
		\centering
		\begin{tikzpicture}
			
			% 第一个坐标轴：左 y 轴 + 蓝线 + 蓝色基准线 + 图例（红线图例也写在这里）
			\begin{axis}[
				xlabel={Length \( S \)},
				ylabel={MSE},
				ymin=0.1, ymax=1,
				ymajorgrids=true,
				axis y line*=left,
				axis x line*=bottom,
				symbolic x coords={1,8,16,32,64,128, 256, 512, 1024},
				xtick=data,
				xticklabel style={rotate=0, font=\small},
				width=11cm,
				height=6cm,
				legend style={
					at={(0.02,0.98)}, anchor=north west,
					font=\scriptsize,
					fill=white,
					draw=gray!30,
					legend cell align=left,
					column sep=1em,
					legend columns=1,
				}
				]
				\addplot[color=red, thick] coordinates {
					(1,0.3874) (8,0.3926) (16,0.4004) (32,0.4074) (64,0.4171) (128,0.4487) (256,0.4716) (512,0.5091) (1024,0.5738)
				};
				\addlegendentry{MSE (KOSS)}
				
				\addplot[red, dashed, thick] coordinates {
					(1,0.5029) (8,0.5029) (16,0.5029) (32,0.5029) (64,0.5029) (128,0.5029) (256,0.5029) (512,0.5029) (1024,0.5029)
				};
				\addlegendentry{MSE (Mamba)}
				
				\addlegendimage{color=myblue, thick}
				\addlegendentry{Throughput (KOSS)}
				
				\addlegendimage{myblue, dashed, thick}
				\addlegendentry{Throughput (Mamba)}
			\end{axis}
			
			% 第二个坐标轴
			\begin{axis}[
				ylabel={Throughput (samples/s)},
				ymin=0, ymax=5000,
				ymajorgrids=true,
				axis y line*=right,
				axis x line=none,
				symbolic x coords={1,8,16,32,64,128, 256, 512, 1024},
				xtick=data,
				width=11cm,
				height=6cm,
				]
				\addplot[color=myblue, thick] coordinates {
					(1,417.504) (8,432.04) (16,469.66) (32,657.04) (64,865.1) (128,1367.46) (256,1804.26) (512,2570.26) (1024,4349.47)
				};
				
				\addplot[myblue, dashed, thick] coordinates {
					(1,2378.04) (8,2378.0) (16,2378.0) (32,2378.0) (64,2378.0) (128,2378.0) (256,2378.0) (512,2378.0) (1024,2378.0)
				};
			\end{axis}
		\end{tikzpicture}
		\caption{
			\textbf{Dynamic Scalability Analysis:} Impact of segment length \( S \) on accuracy and throughput on the ETTm1 dataset. Smaller segments yield lower MSE due to more frequent and fine-grained application of the Kalman-optimal selection mechanism, while larger segments improve throughput via increased parallelism. Dashed lines denote the Mamba baseline. KOSS achieves consistently better accuracy and scales to match or exceed Mamba's throughput under appropriate segment configurations.
		}
		\label{fig:segment-length-ablation}
	\end{figure}
	
	\subsection{Dynamic Scalability Analysis} \label{sec:Segment Length Analysis}
	To evaluate how the segment length \( S \) affects both forecasting accuracy and inference throughput, we conduct a scaling analysis on the ETTm1 dataset. We fix the input sequence length to \( L = 1024 \), and vary the segment length \( S \) across a wide range: \( S \in \{1, 8, 16, 32, 64, 128, 256, 512, 1024\} \).This setting spans the full spectrum from fully recurrent ($S=1$) to fully parallel ($S=L$), enabling analysis of trade-offs between modeling precision and computational throughput. 
	
	As shown in Figure~\ref{fig:segment-length-ablation}, KOSS achieves up to 23\% lower MSE than Mamba at the smallest segment length 
	\( S = 1 \), and consistently outperforms across all segment sizes. Notably, when 
	\( S = 1 \), KOSS effectively degenerates into a linear recurrent structure, which offers the finest granularity of state updates and thus the highest forecasting accuracy under our framework. Moreover, inference throughput increases nearly tenfold as the segment length 
	\( S \) grows. When \( S \) reaches half of the total sequence length, the throughput matches that of Mamba, while KOSS still maintains 14\% higher accuracy. These results demonstrate that KOSS’s segment length parameter effectively balances accuracy and efficiency, enabling flexible deployment tailored to diverse resource constraints.

	\subsection{Real-World Engineering Case Study: SSR Target Trajectory Tracking}
	SSR is a widely used ground-based air traffic surveillance system, where the target replies to an interrogation signal and the ground-based radar system collects sparse detection sequences, known as raw SSR plots, encoding range and azimuth measurements over time. The resulting observation sequences present several challenges for time-series modeling:
	\begin{itemize}
		\item High measurement noise introduces stochastic fluctuations in position estimates;
		\item Irregular and sparse sampling caused by maneuvering targets and radar scan intervals breaks temporal consistency;
		\item Frequent data anomalies or correlated distractors, including spurious echoes or missing detections, cause abrupt discontinuities in the sequence.
	\end{itemize}
	These properties make SSR a natural yet challenging testbed for evaluating the robustness and generalizability of time-series models. The proposed KOSS model integrates context-aware selectivity with learnable state dynamics based on the Kalman optimality principle. It provides a principled framework for filtering, smoothing, and extrapolating trajectories under noisy, correlated distractors and partially observed conditions.

	In this case study, we train KOSS using semi-physical data generated by adding controlled noise to Automatic Dependent Surveillance-Broadcast (ADS-B)~\cite{zhang2011study} flight tracks, simulating measurement uncertainty inherent to SSR systems. For evaluation, we collect real-world SSR echo data from live air traffic in a field environment to directly assess model robustness and reliability under practical conditions. Detailed data preparation and collection procedures are described in Appendix~\ref{appendix:SSR Case Study}.
	
	\begin{figure}[h]
		\centering
		\subfigure[Classical Kalman Filter]{\includegraphics[width=0.23\textwidth]{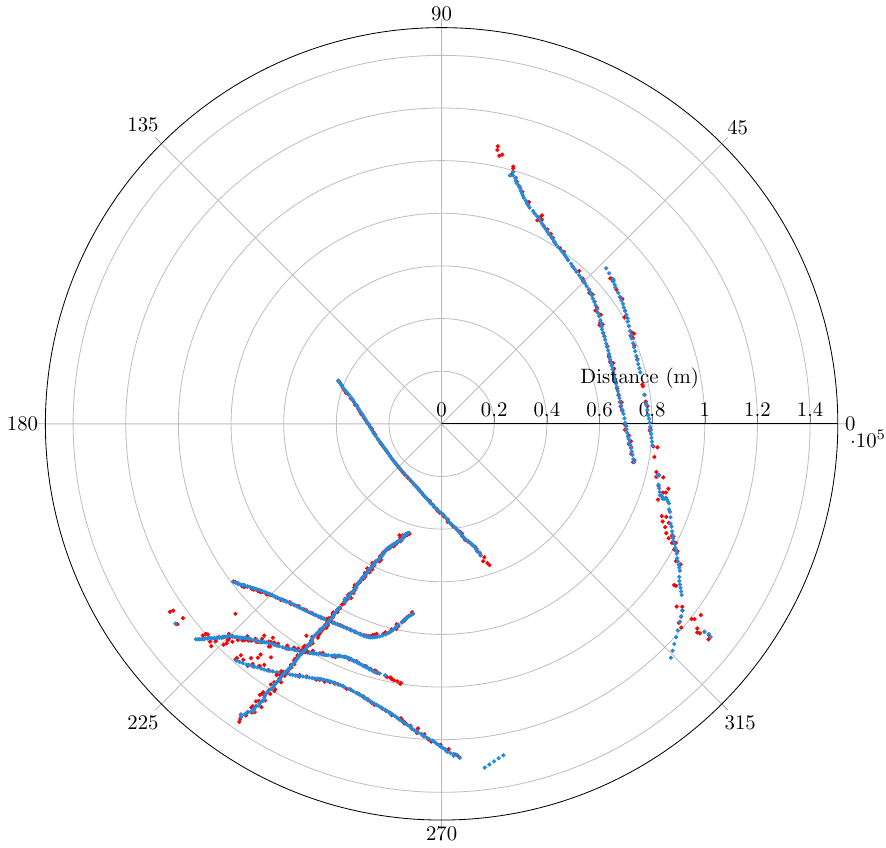}\label{fig:ssr_kalman}}
		\subfigure[Transformer]{\includegraphics[width=0.23\textwidth]{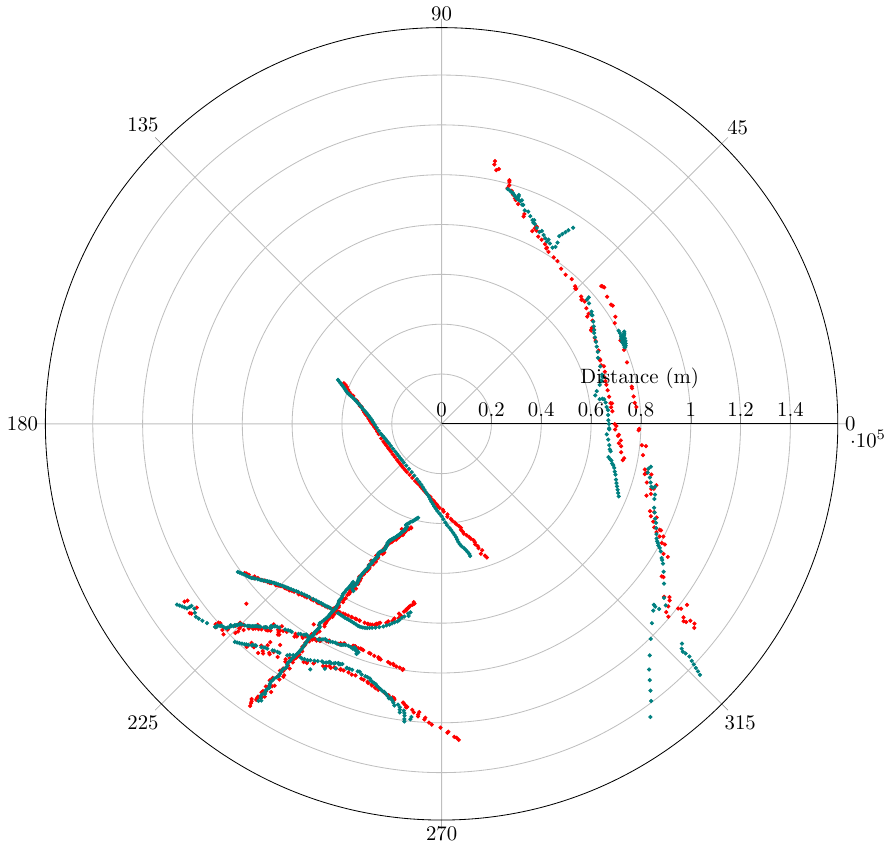}\label{fig:ssr_transformer}}
		\subfigure[Mamba]{\includegraphics[width=0.23\textwidth]{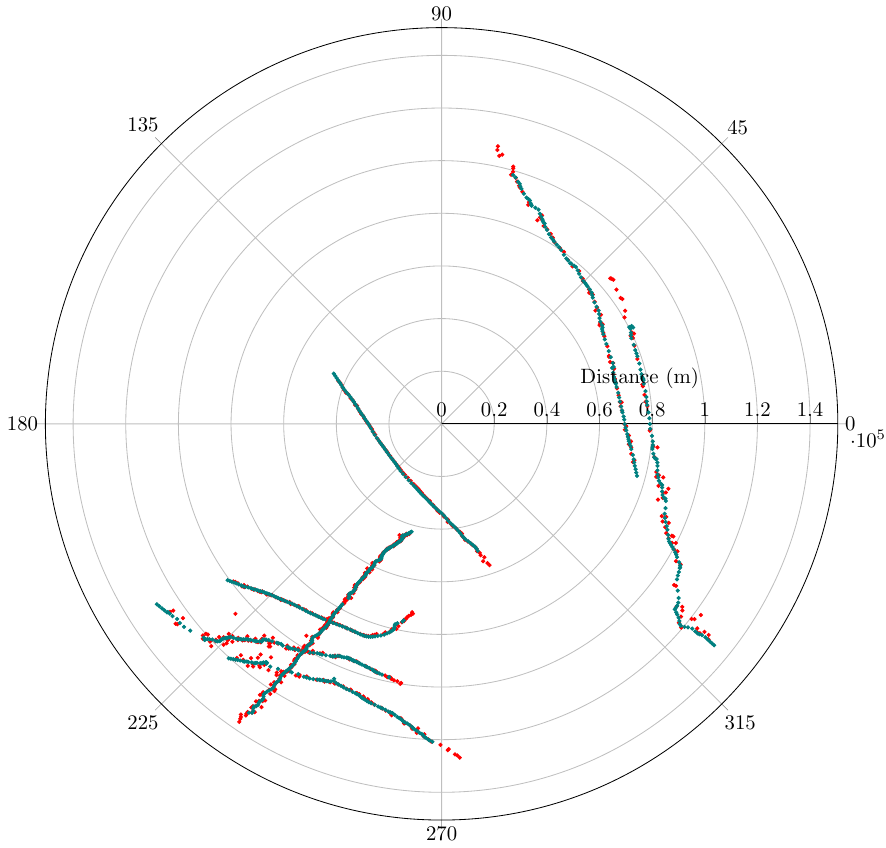}\label{fig:ssr_mamba}}
		\subfigure[KOSS (Ours)]{\includegraphics[width=0.23\textwidth]{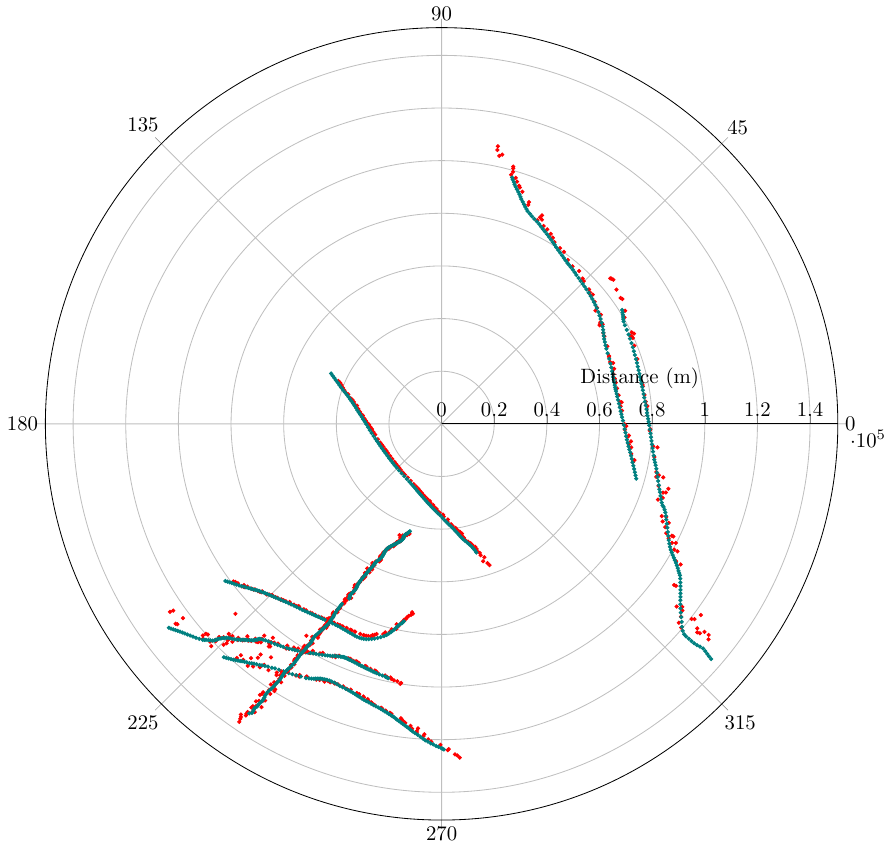}\label{fig:ssr_koss}}
		\caption{\textbf{Trajectory Estimation on Real Raw SSR Plots.}
			Each panel presents the predicted trajectory produced by a specific model from real-world SSR plots collected from 8 live air traffic targets.
			\textcolor{red}{Red}: SSR measurements;
			\textcolor{myblue}{Blue}: Classical Kalman Filter outputs;
			\textcolor{myteal}{Teal}: Deep model (Transformer/Mamba/KOSS)+Kalman Filter outputs.
		}
	\end{figure}
	
	Figure~\ref{fig:ssr_kalman}-\ref{fig:ssr_koss} show a representative trajectory estimation result comparing the classical Kalman Filter, Transformer, Mamba, and our proposed KOSS model. The classical Kalman Filter suffers from trajectory divergence due to fixed dynamics assumptions, often producing tracks that “fly off” from the true path. The Transformer fails to form coherent trajectories. The Mamba model better filters out noise through linear recurrence combined with an input-dependent selectivity mechanism but still struggles to maintain a continuous and stable trajectory during complex maneuvers. In contrast, KOSS achieves smooth and consistent trajectories closely following the true target path even under high noise and irregular observations, demonstrating superior robustness in real-world radar scenarios.

	\subsection{Theoretic Supplement: Stability and Frequency Behavior}  \label{sec:Model-Theoretic-Supplement}
	
	\subsubsection{Stability Analysis of Kalman Gain Dynamics}  \label{sec:convergence-of-Kalman-Gain}
	We empirically examine whether the Kalman gain $K(t)$ stabilizes over time, which supports our modeling assumption $\dot{K}(t) \approx 0$ in Section~\ref{sec:Kalman-Optimal Continuous SSMs}. This assumption plays a critical role in simplifying the state-space model~\eqref{eq:continuous-ssm} without compromising modeling expressiveness.
	
	To validate this, we simulate a controlled linear system under different initial conditions and track the evolution of $K(t)$ over time. We then compare these time-varying gains $K(t)$ with the steady-state solution $K_\infty$ obtained by solving the continuous-time algebraic Riccati equation (CARE)~\eqref{eq:Riccati}, assessing the convergence behavior of the Kalman gain towards its theoretical limit. Detailed system setup and simulation conditions are provided in~\ref{appendix:kalman-convergence}.
	
\begin{figure}[ht]
		\centering
		\includegraphics[width=0.95\linewidth]{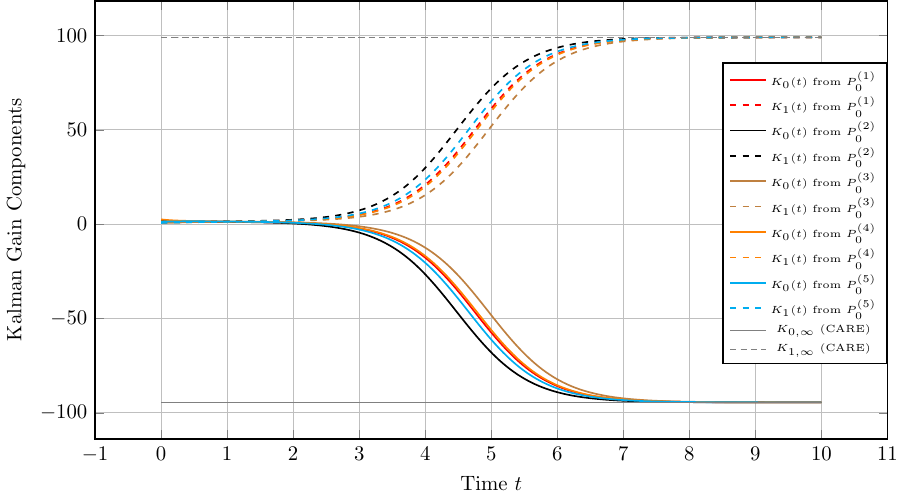}
		\caption{\textbf{Convergence of Kalman Gain from Different Initializations.} Convergence behavior of both components $K_0(t)$ and $K_1(t)$ of the Kalman gain under five distinct initializations $P_0^{(i)}$. Each solid line represents the evolution of $K_0(t)$, while the corresponding dashed line of the same color indicates $K_1(t)$. All trajectories converge to the theoretical steady-state solution $(K_{0,\infty}, K_{1,\infty})$ given by the algebraic Riccati equation.}
		\label{fig:kalman-gain-convergence}
	\end{figure}

	As illustrated in Figure~\ref{fig:kalman-gain-convergence}, the gain components $K_0(t)$ and $K_1(t)$—derived from different initializations—all converge to the same limiting value $K_\infty = [K_{0,\infty}, K_{1,\infty}]$ within a short time horizon. This convergence behavior empirically validates the theoretical result that, under standard observability conditions, $P(t)$ and hence $K(t)$ stabilize over time.
	
	The results support the approximation $\dot{K}(t) \approx 0$ by showing that $K(t)$ rapidly converges to a steady value regardless of its initialization. In practice, ignoring $\dot{K}(t)$ leads to a significant simplification in implementation without compromising modeling fidelity, particularly for long-sequence modeling tasks where temporal stationarity of $K(t)$ is expected.

	\subsubsection{Spectral Response of the SDU Module} \label{sec:frequency-response-of-sdu}
	We further investigate the spectral behavior of the Spectral Differentiation Unit (SDU), a core component responsible for stable derivative estimation in our framework. Unlike standard finite difference methods, SDU performs global differentiation in the Fourier domain, which explicitly enables selective frequency emphasis and improved noise robustness.
	
\begin{figure}[ht]
		\centering
		\includegraphics[width=0.95\linewidth]{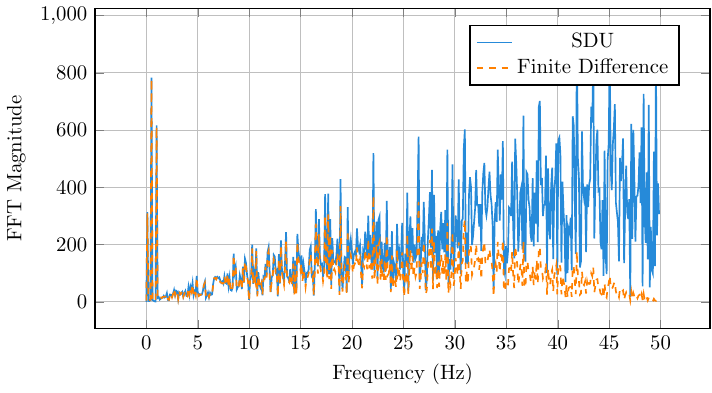}
		\caption{\textbf{Spectral Response of the SDU.} The proposed Spectral Differentiation Unit (SDU) preserves low-to-medium frequency components while attenuating high-frequency noise, demonstrating a more stable and selective differentiation behavior compared to standard finite difference methods.}
		\label{fig:sdu-response}
	\end{figure}

	To assess this, we apply both SDU and a standard central finite difference operator to a mixed-frequency signal contaminated with random noise. We then compare the magnitude spectra of the resulting derivative estimates using the fast Fourier transform (FFT). As shown in Figure~\ref{fig:sdu-response},the SDU exhibits stronger responses to low and medium frequencies while effectively suppressing high-frequency components dominated by noise. In contrast, the finite difference method produces uniformly high responses across all frequencies, reflecting its sensitivity to noise and lack of global context. 
	
	This result highlights SDU’s frequency-selective differentiation property, acting effectively as a global low-pass differentiator. Such behavior improves stability and estimation fidelity, particularly for long sequence modeling tasks involving noisy inputs. Details of the experimental setup are provided in~\ref{appendix:sdu-frequency-response}.
	
\section{Conclusion} \label{sec:Conclusion}
	This work introduces \textbf{KOSS}, a Kalman-Optimal Selective State Space model that rethinks information selection in sequence modeling through the lens of estimation theory. Instead of relying on selection mechanisms driven by simple, ad hoc input-conditioned mappings, KOSS grounds selection in the Kalman optimality principle, yielding a closed-loop, context-aware mechanism.
	
	This formulation offers a unified and efficient modeling framework that bridges principled state estimation with expressive deep learning. It not only challenges the dominant input-only selection mechanisms paradigm but also provides a theoretical lens through which we can interpret the gating mechanisms of classical recurrent architectures such as RNNs and LSTMs—offering a unified explanation for their memory retention and forgetting behaviors.
	
	Extensive experiments demonstrate that KOSS consistently outperforms state-of-the-art baselines in both stability and accuracy across synthetic and real-world scenarios. This includes a field test using raw SSR radar plots, where KOSS achieves reliable trajectory estimation despite measurement noise and temporal sparsity. Furthermore, its segment-wise architecture achieves scalability on par with Mamba while preserving modeling fidelity. Looking ahead, we plan to extend KOSS to broader modalities such as speech, genomics, and multimodal reasoning, where context-aware information integration plays a pivotal role.

	\bibliographystyle{elsarticle-harv} % 使用作者-年份样式
	\bibliography{elsarticle-template-num-names}

	\appendix
	\section{Related Work}
	\subsection{Selection Mechanism} \label{sec:Appendix_Selection_Mechanism}
	Selection mechanisms have emerged as a central design principle in recent state space models, particularly in the development of selective SSMs such as Mamba~(\cite{gu2023mamba}). These mechanisms aim to improve modeling efficiency by dynamically identifying and retaining task-relevant information over long sequences, effectively compressing the state without sacrificing representational power.
	
	\paragraph{Relation to Existing Concepts} 
	Selection mechanisms are conceptually related to various prior ideas, including \textbf{gating}, \textbf{hypernetworks}, and \textbf{data-dependent parameterization}:
	
	\begin{itemize}
		\item \textbf{Gating} originally referred to the signal control logic in RNNs such as LSTM~(\cite{hochreiter1997long}) and GRU~(\cite{cho2014learning}), where gating functions regulate memory updates. While the term has since been generalized to denote any form of multiplicative interaction, such as in gated convolutional architectures~(\cite{mehta2022long, hua2022transformer}), this generalization often departs from the original notion of temporal signal control.
		
		\item \textbf{Hypernetworks}~(\cite{ha2016hypernetworks}) refer to networks whose parameters are dynamically generated by auxiliary neural networks. When used in sequence modeling, they allow recurrent or convolutional parameters to vary based on the input signal, enabling a degree of data adaptivity.
		
		\item \textbf{Data-dependence}~(\cite{poli2023data}) broadly refers to models where parameters are conditioned on input data. In this sense, both gating and hypernetworks fall under this larger umbrella.
	\end{itemize}
	
	Although selection mechanisms share similarities with the above concepts, they constitute a \textit{distinct category}. In particular, they are specifically designed to route, filter, or suppress sequence-level information in an input- or state-dependent manner, facilitating efficient modeling over long horizons. In selective SSMs, this is typically achieved by parameterizing SSM components (e.g., $\Delta$, $A$, $B$, $C$) as functions of the input at each time step.
	
	\paragraph{From Implicit to Explicit Selection} 
	Early structured SSMs like S4~(\cite{gu2021combining}) encode fixed inductive biases via learned structured dynamics, providing an implicit form of selection through signal propagation. Later models such as Mamba~(\cite{gu2023mamba}) introduced \textbf{explicit selection}, where parameters of the state space (e.g., $\Delta$, $A$, $B$, $C$) are directly conditioned on the current input, enabling the model to dynamically emphasize or ignore input features at each step.
	
	\paragraph{Semantic Clarification}
	Mamba authors~(\cite{gu2023mamba}) argue that although selection mechanisms can be loosely categorized as gating, hypernetworking, or data dependence, this view is too broad to be meaningful. Instead, they advocate using the term \textbf{selection} to denote mechanisms that operate specifically over the \textit{sequence dimension}, enabling long-range control and adaptive memory compression. Such mechanisms often align more closely with classical RNN gating when viewed through the lens of state space systems and dynamic discretization~(\cite{funahashi1993approximation, tallec2018can}).
	
	\paragraph{Scope and Relevance}
	The selection principle underpins the recent progress in linear-time sequence models. It enables the design of architectures like Mamba and our proposed KOSS model, where selection not only depends on input features but also incorporates latent state dynamics via innovation signals. A more comprehensive overview of selection variants and their computational forms is provided in Section~(\ref{sec:selective_mechanisms}).

	\subsection{From LTI to LTV Systems} \label{appendix:Appendix_LTV}
	The most significant characteristic in the progression from S4 to S6 is the ability to efficiently select information in an input-dependent manner. This introduces a fundamental shift: the parameters \((\Delta, A, B, C)\) and \((\bar{A}, \bar{B})\) vary with the input, thereby forming a Linear Time-Varying (LTV) system. This time-varying nature has two important implications:
	\begin{itemize}
		\item \textbf{Theoretical Perspective:} a fundamental problem of sequence modeling is compressing context into a smaller state(~\cite{gu2023mamba}). From this perspective, recurrent models are efficient because they have a finite state, implying constant-time inference and linear-time training. However, their effectiveness is limited by how well this state  has compressed the context. In traditional linear time-invariant (LTI) systems, the state evolution follows a fixed propagation path determined by static transition operators, which limits their capacity to adaptively compress context under dynamically changing patterns. In contrast, linear time-varying (LTV) systems such as S6 employ input-dependent dynamic parameterization, enabling input- and position-sensitive pathways for information injection and extraction. This gives rise to a form of data-dependent routing similar to that of Transformers, albeit through a fundamentally different mechanism—where Transformers compute pairwise similarity between token representations over the entire sequence. Nevertheless, the resulting structure can be viewed as an intermediate paradigm between the hidden state evolution of traditional RNNs and the fully connected attention in Transformers.
		
		\item \textbf{Computation Perspective:} The LTV structure poses two key challenges: the inherently sequential nature of linear recurrence and high memory consumption. To address these limitations, recent implementations leverage a combination of kernel fusion, parallel scan, and recomputation. The core idea is to compute the model through a scan rather than through linear recurrence or convolution, avoiding materialization of the expanded state and thereby reducing I/O traffic across GPU memory hierarchy levels. The resulting implementation achieves superior efficiency compared to prior methods, scaling linearly with sequence length in theory and delivering up to 3× speedup on A100 GPUs in practice.
	\end{itemize}

	\section{Connection Between Innovation-Driven Selectivity and Gating Mechanisms}
	\label{appendix:rnn-connection}
	
	We examine the theoretical connection between the proposed innovation-driven selectivity in KOSS and the gating behaviors commonly found in  classical recurrent neural networks (RNNs, LSTMs, GRUs). At a functional level, both approaches aim to regulate the flow of information over time—deciding what to retain, update, or discard during sequence processing. While RNNs implement this through specialized gates (e.g., input gate, forget gate) trained end-to-end, KOSS derives an analogous form of information selection by minimizing latent state uncertainty, resulting in an adaptive Kalman gain that modulates information propagation.

	\begin{itemize}
		\item The input and forget gates in LSTM/GRU can be interpreted as input and state-driven information control factors: the input gate heuristically determines which aspects of the current input should influence the hidden state, while the forget gate controls the retention of historical information.
		\item KOSS explicitly models this control by computing a Kalman-optimal gain $K$, establishing a closed-loop relationship between the current input and latent state estimates. The Kalman gain adaptively weights the contributions of input content and state context for selective information propagation.
	\end{itemize}
	
	This correspondence highlights that gating in RNNs and selectivity in KOSS can be interpreted as different realizations of the same underlying principle: dynamically adjusting information propagation to optimize predictive fidelity. By grounding this mechanism in Kalman filtering, KOSS offers a unified theoretical lens that helps explain and generalize the function of memory gates in classical sequence models.

	\section{Kalman Gain as a Prototype for Dynamic Selectivity in Deep Models} \label{appendix:Gain is a Selective}
	In classical filtering theory, the Kalman gain \( K(t) \) acts as a dynamic weighting factor that determines how prior state estimates and new observations are combined to produce a new state~\cite{kalman1960new}. It is formally derived as the solution to the Riccati differential equation, and its value depends on the evolving uncertainty in the internal system state (captured by the prior covariance \( P(t) \)) and the noise characteristics of the observation process (captured by the measurement covariance \( R \))~\cite{tahirovic2025optimal}. This time-varying gain governs the extent to which incoming measurements correct the state estimate, ensuring minimum mean-squared error under Gaussian noise assumptions.
	
	This correction process is essentially a trade-off between the historical state information of the system and the incoming input information. In the context of deep learning, we interpret the observation as the current input sequence, and the prior state estimate as the latent representation of the system's history. From this perspective, the Kalman gain plays a role analogous to a dynamic selection factor that balances the contribution of contextual knowledge (from the model state) and content information (from the input) in generating the updated representation. Therefore, the Kalman gain can be viewed as a principled prototype for content-aware and context-sensitive information selection. This perspective justifies the broader use of dynamic selection strategies in sequence models, especially when aiming to balance stability and adaptability in long-range modeling.

	\section{Justifying the Approximation $\dot{K}(t) \approx 0$} \label{appendix:kalman-derivative}
	As shown in Equation~\ref{eq:continuous-ssm}, the continuous-time Kalman-optimal state-space model involves the time derivative of the Kalman gain $\dot{K}(t)$. However, to simplify implementation and enhance stability, we argue that it is reasonable to treat this term as approximately zero from both theoretical and practical perspectives.
	
	\begin{itemize}
		\item \textbf{Theoretical perspective.} In addition to deriving Kalman gain $K(t)$ by minimizing the state estimation error, one can also obtain it by solving the Riccati equation~\cite{tahirovic2025optimal}. According to this formulation, when the system is observable and controllable, the Riccati equation admits a unique stable covariance solution $P^*$, which implies that covariance $P(t)$ eventually stabilizes as time $t$ increases. Consequently, the gain $K(t)$ also converges to a steady value $K^*$, and its time derivative $\dot{K}(t)$ becomes negligible~\cite{grigorian2022properties}. In long-term time series forecasting tasks, the model gradually adapts to the input distribution and noise characteristics, leading to a stabilized weighting between prior and new information~\cite{de2018long,chen2019forecasting}. As a result, the approximation $\dot{K}(t) \approx 0$ is theoretically justified. 
		
		\item \textbf{Practical perspective.} From a modeling standpoint, explicitly estimating or incorporating $\dot{K}(t)$ poses significant challenges~\cite{dahan2024uncertainty,shen2025kalmanformer}. First, the derivative of the Kalman gain is highly sensitive to input perturbations, which may amplify errors during training and result in unstable gradients, oscillations, or optimization failure---especially in neural models~(\cite{dahan2023bayesian}). Moreover, introducing $\dot{K}(t)$ introduces additional dynamic parameters and nonlinear dependencies. This leads to increased computational complexity, more difficult optimization, and potential non-stationarity~\cite{revach2022kalmannet}. These challenges can ultimately interfere with the core learning objective and degrade the model’s generalization ability. In contrast, approximating $\dot{K}(t) \approx 0$ simplifies the model architecture, reduces training difficulty, and significantly improves optimization stability and efficiency. This simplification is particularly beneficial for long-sequence modeling~\cite{gawlikowski2023survey}.
	\end{itemize}

	\section{Spectral Differentiation Unit (SDU)}\label{appendix:sdu-details}
	This appendix expands the derivation of the Spectral Differentiation Unit (SDU) introduced in Section~(\ref{sec:SDU}), providing numerical implementation details and Fourier frequency handling techniques.

	\subsection{Mathematical Derivation}
	
	Given a real-valued input sequence $x = \{x_0, x_1, \dots, x_{N-1}\}$ sampled at uniform intervals $\Delta t$, we aim to estimate the time derivative $\frac{dx}{dt}$.
	
	We apply the Discrete Fourier Transform (DFT) to obtain the frequency domain representation:
	
	\[
	X_k = \sum_{n=0}^{N-1} x_n \cdot e^{-j 2\pi kn / N}
	\]
	
	According to the continuous-time identity $\mathcal{F}[x'(t)] = j\omega \cdot \mathcal{F}[x(t)]$, we estimate the frequency-domain derivative as:
	
	\[
	X'_k = j \omega_k \cdot X_k, \quad \omega_k = \frac{2\pi k}{N \Delta t}
	\]
	
	The corresponding time-domain derivative is recovered using the inverse DFT:
	
	\[
	x'_n = \text{IDFT}(X'_k) = \frac{1}{N} \sum_{k=0}^{N-1} j\omega_k X_k \cdot e^{j 2\pi kn / N}
	\]
	
	This computation can be implemented efficiently via Fast Fourier Transform (FFT) and inverse FFT. Let $\texttt{FFT}(x)$ denote the FFT of input $x$ and $\texttt{IFFT}(\cdot)$ the corresponding inverse operation. Then, the SDU procedure is:
	
	\begin{enumerate}
		\item Compute $X = \texttt{FFT}(x)$
		\item Multiply each component $X_k$ by $j\omega_k$
		\item Compute $x' = \texttt{IFFT}(j\omega \cdot X)$
	\end{enumerate}
	
	\subsection{Practical Implementation}
	
	\paragraph{Frequency Vector Construction.}
	The vector $\omega = [\omega_0, \omega_1, \dots, \omega_{N-1}]$ is constructed as:
	
	\[
	\omega_k = 
	\begin{cases}
		\frac{2\pi k}{N \Delta t}, & 0 \le k < N/2 \\
		\frac{2\pi (k - N)}{N \Delta t}, & N/2 \le k < N
	\end{cases}
	\]
	
	This symmetric definition ensures correct handling of positive and negative frequencies in FFT implementations.
	
	\paragraph{Stability Considerations.}
	High-frequency components are often corrupted by noise. To improve numerical stability, we apply a soft spectral mask or frequency truncation:
	
	\[
	\tilde{X}_k = j\omega_k \cdot X_k \cdot \chi(\omega_k), \quad \chi(\omega_k) \in [0,1]
	\]
	
	where $\chi(\omega)$ is a damping function such as:
	
	\[
	\chi(\omega) = \exp\left(-\frac{|\omega|}{\omega_{\text{cut}}} \right)
	\]
	
	or a hard cutoff mask $\chi(\omega) = \mathbb{I}[|\omega| \le \omega_{\max}]$.
	
	\paragraph{Complex Output Handling.}
	Since FFT outputs are generally complex-valued, the resulting $x'_n$ may also contain small imaginary parts due to numerical error. We discard the imaginary component via:
	
	\[
	x'_n \gets \Re(x'_n)
	\]
	
	\subsection{Integration with Kalman-Optimal Architecture}
	
	The derivative sequence $x'_n$ computed via SDU is fed directly into the recurrence relation of our Kalman-Optimal state-space model:
	
	\[
	h_t = A_K h_{t-1} + B_K x_t + K \cdot x'_t
	\]
	
	This integration allows the model to capture input dynamics at fine temporal granularity, with stable and efficient derivative estimation across long sequences.

	\section{Details of Segment-wise Parallel Scan} \label{appendix:segment-scan-details}
	\paragraph{Computation Structure} Given input \(X \in \mathbb{R}^{B \times L \times D}\) with segment length \(S\), we split the sequence into \(M = \lceil L / S \rceil\) segments, denoted by index \(\ell\). Within each segment:
	\[
	\textsc{Scan}(\overline{A}_K^{(\ell)} H^{(\ell-1)},~ \overline{B}_K^{(\ell)} X^{(\ell)} + K^{(\ell)} \delta X^{(\ell)}) \rightarrow H^{(\ell)} \in \mathbb{R}^{B \times S \times D \times N}.
	\]
	For \(\ell=0\), the initial state \(H^{(-1)}\) is set to zeros.
	
	\paragraph{Implementation Notes}
	\begin{itemize}
		\item \textbf{Parallel intra-segment scan:} Using GPU scan primitives or custom CUDA kernels, unrolling or warp-level scan for small \(S\) (e.g., 16/32).
		\item \textbf{Batch × segment parallelism:} Kernels are launched over \(B \times S\) blocks.
		\item \textbf{Latency:} Intra-segment dependency chain is of length \(S\), while inter-segment recurrence scales linearly with \(M=\lceil L/S \rceil\), leading to total latency \(\frac{L}{S} + \log S\).
	\end{itemize}
	
	\paragraph{Efficiency Discussion} 
	when the parameters \( A, B \) in recurrent SSMs are fixed—either through HiPPO parameterization or as learnable constants—the system reduces to a fixed convolution kernel, enabling all \( h_t \) to be generated in \( \mathcal{O}(1) \) depth with full parallelism of \( B \times L \). However, this approach requires explicitly maintaining a hidden state sequence of length \( L \). In contrast, hardware-aware techniques (e.g., block-wise exponential scan) construct a compact summary of past states using prefix scan primitives while preserving recursive consistency, thus avoiding the need to store the entire state sequence and significantly reducing memory cost.
	This, however, comes at the cost of decoupling \( A, B \) from \( h_t \). In our method, \( A, B, K \) are tightly coupled with the input \( x_t \) and hidden state \( h_{t} \), leading to mutual dependence and cycles in the computation graph. We therefore adopt a segment-level parallel scan combined with cross-segment recurrence, tracking only \( M = L/S \) hidden states.

	\section{Experimental Details and Additional Results}
	\subsection{Context-Aware Selective Copying} \label{sec:Appendix Synthetic Tasks}
	Our experiments follow prior work~(\cite{gu2023mamba}), using sequences of length 4096 with a vocabulary size of 16 tokens, including data tokens, white noise tokens, and faint-colored distractor tokens (see Figure~\ref{fig:copy-vs-selective}). Each sequence contains 16 data tokens that the model must memorize. The distractor token interference levels are set between 0\% and 50\% relative to the data tokens. We use 2-layer models with a hidden dimension of \(D = 64\).
	
	Models are trained for 400K steps with a constant learning rate of 0.0001 and a batch size of 64.
	
	\begin{table}[ht]
		\centering
		\caption{Details of benchmark datasets used in our experiments.}
		\begin{tabular}{@{}lcccc@{}}
			\toprule
			\textbf{Dataset} & \textbf{Frequency} & \textbf{\# Features} & \textbf{Length} & \textbf{Time Span} \\ \midrule
			ETTh1      & 1 hour      & 7   & 17,420  & 2016–2017 \\
			ETTh2      & 1 hour      & 7   & 17,420  & 2017–2018 \\
			ETTm1      & 15 minutes  & 7   & 69,680  & 2016–2017 \\
			ETTm2      & 15 minutes  & 7   & 69,680  & 2017–2018 \\
			Exchange   & 1 day       & 8   & 7,588   & 1990–2010 \\
			Weather    & 10 minutes  & 21  & 52,696  & 2020 \\
			Electricity& 1 hour      & 321 & 26,304  & 2012–2014 \\
			ILI        & 7 days      & 7   & 966     & 2002–2020 \\
			Traffic    & 1 hour      & 862 & 17,544  & 2015–2016 \\
			\bottomrule
		\end{tabular}
		\label{tab:dataset-details}
	\end{table}
	\subsection{Main Experiments} \label{sec:Appendix Main Experiments}
	\paragraph{Dataset Details} \label{sec:dataset_details}
	We summarize the datasets used in this study as follows. 
	\textbf{Weather}~\cite{weather_dataset_2024} contains 21~meteorological variables (e.g., temperature and humidity) recorded every 10 min throughout 2020. 
	\textbf{ETT} (Electricity Transformer Temperature)~\cite{ett_dataset_2024} includes four subsets: two hourly-level datasets (ETTh1, ETTh2) and two 15-minute-level datasets (ETTm1, ETTm2). 
	\textbf{Electricity}~\cite{electricity_dataset_2024}, derived from the UCI Machine Learning Repository, records hourly power consumption (kWh) of 321 clients from 2012 to 2014. 
	\textbf{Exchange}~\cite{exchange_dataset_2024} comprises daily exchange rates among eight countries. 
	\textbf{Traffic}~\cite{traffic_dataset_2024} consists of hourly road occupancy rates measured by 862 sensors on San Francisco Bay Area freeways from January 2015 to December 2016. 
	\textbf{Illness} (ILI) dataset~\cite{ili_dataset_2024} tracks the weekly number of influenza-like illness patients in the United States. 
	Table~\ref{tab:dataset-details} summarizes the statistical properties of all nine benchmark datasets. 
	All datasets are divided into training, validation, and test subsets with a ratio of 7:1:2.
	
	\paragraph{Implementation Details} 
	We use the Adam optimizer with no weight decay. All models are trained at constant learning rates ranging from \(1 \times 10^{-3}\) to \(1 \times 10^{-2}\). The batch size is set to 32 by default, depending on available GPU memory. We observe that increasing the batch size up to 256 does not degrade performance, while offering faster training on GPUs with larger memory or in multi-GPU setups. The training runs for 15 epochs without early stopping, and the checkpoint with the lowest validation loss is used for final evaluation. We report mean squared error (MSE) and mean absolute error (MAE) as the performance metrics. Each experiment is repeated five times, and the average results are reported. All models are implemented in PyTorch 2.1.0 with Python 3.11, and trained on NVIDIA V100 GPUs (16GB and 32GB variants).

	\subsection{SSR Case Study: Data and Preprocessing Details} \label{appendix:SSR Case Study}
	
	\paragraph{Data Acquisition and Noise Simulation}
	The semi-physical training data is sourced from the OpenSky Network ADS-B logs collected on June 15, 2025, between 10:00–10:15 AM, covering a 300km region around Tokyo Haneda, Narita, and Incheon Airports. Each airport provides 15 minutes of flight data at a 5-second refresh rate, capturing 503 targets across various motion modes (descent, climb, turn, cruise, etc.). Each record includes 17 attributes; we use UTC timestamp, longitude, latitude, and altitude as input features. A detailed summary of the ADS-B-based semi-physical dataset is presented in Table~\ref{tab:ADS_B-dataset-details}.

	To emulate SSR noise patterns, Gaussian noise with an SNR of 33dB is added to each sample. This noise level corresponds to typical SSR measurement conditions.
	\[
	X_i = X_i + \mathcal{N}(0, \sigma_i^2 / 10^{\frac{33}{10}} )
	\]
	where \(X_i\) is the sample and \(\sigma_i\) is its standard deviation. 
	
	This setup enables a rigorous evaluation of model performance under conditions closely approximating actual radar operation
	
	\paragraph{Normalization and Reverse Transformation}
	
	All input data is standardized to zero-mean, unit-variance using:
	\[
		y = \frac{x - \bar{x}}{\sigma},
	\]
	where \(x\) denotes the original sample vector; \(\bar{x}\) represents the sample mean; \(\sigma\) is the sample standard deviation; and \(y\) denotes the standardized sample vector.
	 During inference, predictions are reverse-transformed:
	\[
		\hat{x} = \sigma \hat{y} + \bar{x}
	\]

	\begin{table}[ht]
		\centering
		\caption{Summary of the semi-physical ADS-B dataset used for training.}
		\small
		\begin{tabular}{@{}cccccc@{}}
			\toprule
			\textbf{Targets} & \textbf{Frequency} & \textbf{\# Features} & \textbf{Time Steps} & \textbf{Time Span} \\ \midrule
			503 & 5 seconds & 4 (UTC, Lon, Lat, Alt) & 166,110 & 2025-06-15, 10:00–10:15 \\ 
			\bottomrule
		\end{tabular}
		\label{tab:ADS_B-dataset-details}
	\end{table}
	
	\paragraph{Field-Collected SSR Data}
	For real-world evaluation, we use field-collected SSR plots from a deployed radar system (with a peak transmit power of 6 kW) during live tracking operations. The collected data comprise eight targets and exhibit irregular sampling intervals, high measurement noise, and spurious or missing returns. We use these unprocessed plot sequences as input to rigorously assess model robustness and generalization in authentic operational environments.

	\begin{figure}[h]
		\centering
		\includegraphics[width=0.85\textwidth]{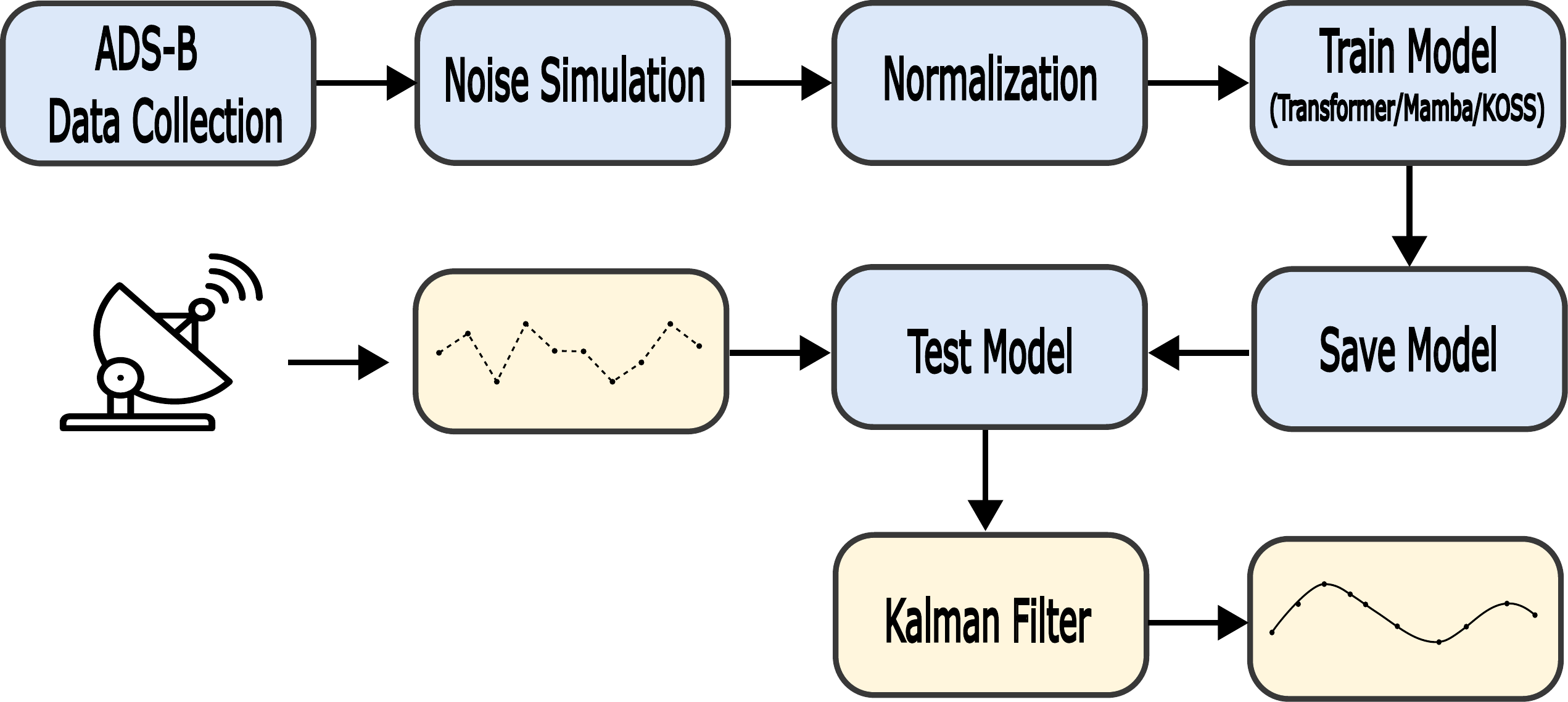}
		\caption{\textbf{SSR Trajectory Modeling Pipeline.}  The upper branch shows training using ADS-B data with noise simulation and normalization; the lower branch shows testing on real raw SSR plots using trained models, followed by trajectory prediction and comparison.
		}
		\label{fig:ssr_flow}
	\end{figure}

   \paragraph{Processing Workflow}
	The full pipeline, from semi-physical data generation to field-test evaluation, is illustrated in Figure~\ref{fig:ssr_flow}, which separates training and testing into two branches:
	
	\begin{itemize}
		\item \textbf{Training Path (Upper Branch)}: ADS-B flight trajectories are collected from the OpenSky Network and passed through a noise simulation module to emulate SSR raw plots characteristics (e.g., low SNR, sparse sampling). The data is then normalized and used to train deep sequence models including Transformer, Mamba, and KOSS. After training, model parameters are saved for downstream testing.
		
		\item \textbf{Testing Path (Lower Branch)}: Real SSR raw point detections data is collected using a field-deployed radar. These point detections exhibit strong noise and temporal sparsity. The trained models are loaded and applied directly to these time series for trajectory estimation. The predicted outputs are then smoothed and visualized for qualitative comparison.
	\end{itemize}
	
	This structured workflow mirrors the real-world deployment scenario where models are trained on controlled data and validated on noisy, unlabeled measurements.

	\subsection{Detailed Setup for Kalman Gain Convergence Experiment}
	\label{appendix:kalman-convergence}
	
	We simulate a low-dimensional continuous-time linear system to examine the convergence behavior of the Kalman gain $K(t)$ under different initial covariance conditions. The system dynamics follow the Riccati differential equation:
	\begin{equation}
		\frac{dP}{dt} = A P + P A^\top - P B R^{-1} B^\top P + Q,
		\label{eq:Riccati}
	\end{equation}
	with fixed system parameters:
	\[
	A = \begin{bmatrix} 0.9 & 0.0 \\ 0.0 & 0.95 \end{bmatrix}, \quad
	B = \begin{bmatrix} 1.0 \\ 1.0 \end{bmatrix}, \quad
	Q = I, \quad
	R = [1.0].
	\]
	
	We initialize $P(0)$ with five distinct positive definite matrices, ranging from the identity to perturbed low-rank forms. At each time step, the corresponding Kalman gain is computed as
	\[
	K(t) = R^{-1} B^\top P(t),
	\]
	and its components $K_0(t)$ and $K_1(t)$ are tracked over time. For reference, the steady-state solution $K_\infty$ is obtained by solving the continuous-time algebraic Riccati equation (CARE):
	\[
	A P + P A^\top - P B R^{-1} B^\top P + Q = 0.
	\]
	
	This experiment supports the classical result that, for stable observable systems, the Riccati solution $P(t)$ converges to a unique positive-definite solution, and thus $K(t)$ converges to a steady-state gain $K_\infty$~\cite{lancaster1995algebraic}.

	\subsection{Detailed Setup for SDU Frequency Response Experiment} \label{appendix:sdu-frequency-response}
	
	To evaluate the frequency-domain behavior of the proposed Spectral Differentiation Unit (SDU), we constructed a synthetic signal comprising a mix of low, medium, and high frequency components with additive Gaussian noise. The input sequence takes the form:
	\[
	x(t) = \sin(2\pi f_1 t) + 0.5 \sin(2\pi f_2 t) + 0.2 \sin(2\pi f_3 t) + \text{noise},
	\]
	where \( f_1 = 0.1 \), \( f_2 = 0.5 \), and \( f_3 = 1.0 \). The noise is standard normal and independent across time.
	
	We compute the time derivative of $x(t)$ using two methods:
	(1) The SDU, which performs spectral-domain differentiation via FFT;
	(2) A classical finite difference operator using the central difference kernel \([1, 0, -1] / 2\Delta t\).
	
	We then apply FFT to both derivative estimates and plot their respective magnitude spectra. These resulting curves directly reveal the distinct frequency-selective behaviors of the SDU.

\end{document}